\documentclass[runningheads]{llncs}
\usepackage[T1]{fontenc}
\usepackage{graphicx}
\usepackage{times}
\usepackage{epsfig}
\usepackage{graphicx}
\usepackage{amsmath}
\usepackage{amssymb}

\usepackage{tikz}
\usepackage{comment}
\usepackage{color}
\usepackage[utf8]{inputenc}
\usepackage{subcaption}
\usepackage{amsfonts} 
\usepackage{tabularx}
\usepackage{amsfonts}
\usepackage[mathscr]{euscript}

\begin{document}
\title{Self-supervised learning of object pose estimation using keypoint prediction}
%
%

\author{Zahra Gharaee \inst{1,2}\orcidID{0000-0003-0140-0025} \and
Felix J\"aremo Lawin\inst{1}\orcidID{0000-0003-4830-0443} \and
Per-Erik Forss\'en\inst{1}\orcidID{0000-0002-5698-5983}}
\authorrunning{F. Author et al.}
%
\institute{
	Computer Vision Laboratory (CVL), Department of Electrical Engineering, University of Link\"oping, Link\"oping, Sweden\\
\email{felix.lawin@gmail.com,per-erik.forssen@liu.se}\\
\and
Vision and Image Processing Laboratory (VIP), Systems Design Engineering, University of Waterloo, Waterloo, Canada\\
\email{zahra.gharaee@uwaterloo.ca}}
\maketitle              

\begin{abstract}
This paper describes recent developments in object specific pose and shape prediction from single images.
The main contribution is a new approach to camera pose prediction by self-supervised learning of keypoints corresponding to locations on a category specific deformable shape. We designed a network to generate a proxy ground-truth heatmap from a set of keypoints distributed all over the category-specific mean shape, where each is represented by a unique color on a labeled texture. The proxy ground-truth heatmap is used to train a deep keypoint prediction network, which can be used in online inference. The proposed approach to camera pose prediction show significant improvements when compared with state-of-the-art methods.

Our approach to camera pose prediction is used to infer 3D objects from 2D image frames of video sequences online. To train the reconstruction model, it receives only a silhouette mask from a single frame of a video sequence in every training step and a category-specific mean object shape.
We conducted experiments using three different datasets representing the bird category: the CUB \cite{wah2011caltech} image dataset, YouTubeVos and the Davis video datasets. The network is trained on the CUB dataset and tested on all three datasets. The online experiments are demonstrated on YouTubeVos and Davis \cite{xu2018youtube} video sequences using a network trained on the CUB training set.

\keywords{Self-supervised learning \and 3D shape prediction  \and camera pose estimation \and keypoint prediction}
\end{abstract}
\section{Introduction}
\vspace{-0.8em}

People looking at 2D images of humans, animals, and objects, can infer much more information than just size, color, and light. This implicit information relates to the 3D structures of all entities shown by a 2D image. This ability is not only limited to inferring the 3D shape of an object and the camera pose capturing it but it also extends to imagining the object from novel views.

More fascinating is that as humans, we do not rely on an already seen image of a particular object instance. We rather apply a categorical inference using a prototypical member of a kind based on our previous experience of observing various members of a category like e.g. human \cite{bulthoff1998top}.

Inspired by this human ability to extract 3D information from 2D images, there has been a surge in research proposing models to infer 3D shapes from 2D images as well as predicting the camera pose capturing it \cite{Niemeyer2020CVPR,mesh_rcnn2019,Haozhe2019,cmr2018,ucmr2020}. 3D object reconstruction has recently made innovations in applications like e-commerce, computer graphics, special effects, augmented reality, visual localization, and computational photography.

This paper introduces a novel model for self-supervised keypoints based learning of the pose, which is used in inferring 3D objects reconstructed from 2D images and video sequences.

\vspace{-0.7em}
\section{Related work} 
\vspace{-0.7em}

Learning-based approaches to 3D object reconstruction map 2D images to various kinds of output representations, including voxels \cite{3D-R2N2,Haozhe2019,Gernot2017}, which are limited to relatively small size due to memory constraints, points \cite{Haoqiang2017,Jiang2018,Hugues2019}, which require extensive post-processing to extract connectivity information, meshes \cite{Hiroharu2018,Jhony2018,Nanyang2018,mesh_rcnn2019}, which are more applicable to deformable objects, and implicit representations \cite{Zhiqin2019,Kyle2019,Mateusz2019,Shunsuke2019,Niemeyer2020CVPR}. Our proposed approach is in the category of 3D mesh reconstruction of the objects that appear in 2D images.

\vspace{-0.5em}
\subsection{Mesh reconstruction challenges}
\vspace{-0.7em}

The main objectives of most approaches to shape reconstruction is to extract a 3D model of the object, its texture as well as the camera parameters. However, many challenges are facing an object's 3D shape reconstruction. We list a few of them in the following.

One challenging problem appears in rigidity of an entity's shape reconstructed from 2D images, since it is much easier to infer 3D shapes of rigid objects \cite{Kar2015,Lin2019} like cups, compared to non-rigid entities like a human body \cite{Angjoo2019,Jason2019}, faces \cite{Shangzhe2020}, hands \cite{Jonathan2014,Sameh2015}, quadruped animals \cite{Silvia2017} and birds \cite{cmr2018,ucmr2020}. Especially in motion, the deformation of non-rigid objects results in much larger variations of the shapes.

Deformation of dynamic real-world objects preserves topology, and nearby points are more likely to deform similarly compared to more distant points \cite{DEMEA2020}. As an example, the 3D structure of a human body is restricted to ground-truth constraints from skeleton postures, which results in local rigidity, the links connecting consecutive joints are rigid. As-Rigid-As-Possible \cite{ARAP2007} applies smoothness and local rigidity, based on a set of hand-crafted priors in the reconstructions. DEMEA \cite{DEMEA2020} also reasons about the local rigidity of meshes to enable higher-quality results for highly deformable objects.

A next challenge is whether 2D images are captured in a constrained \cite{3D-R2N2,Angjoo2019,Jason2019} or more naturalistic environment \cite{cmr2018,ucmr2020,Li2020OnlineAF}. This becomes essential when generalizing the trained models to unseen images or videos of the objects in realistic environments. 

\vspace{-0.5em}
\subsection{Amount of supervision}
\vspace{-0.5em}

One of the most challenging problems is the amount of supervision needed to train a model. This has been studied extensively by different approaches to 3D reconstruction of objects from 2D images. In 3D supervision, ground-truth 3D shape is used as a supervisory signal to learn reconstruction from single images with voxel-based representations \cite{3D-R2N2,Haozhe2019}, point-based representation \cite{Haoqiang2017,Jiang2018} and mesh-based representations \cite{Hiroharu2018,Jhony2018,Nanyang2018,mesh_rcnn2019} as outputs of the network.

In 2D supervision, depth maps or multiple views of the same object are used to recover 3D structure through known viewpoints \cite{Xinchen2016,kar2017,Shubham2017} or unknown viewpoints \cite{Matheus2017,Eldar2018,Shubham2018}. Although applying multiple viewpoints is less supervised compared to the ground-truth 3D supervision, it still requires having access to the multiple views of silhouettes, images, and depth since it cannot reconstruct unseen parts of the object.

To address this challenge, other methods \cite{Cashman2013,Kar2015,cmr2018} replaced the need for 3D data by only fitting a 3D model to 2D-keypoints and a segmentation mask. Their model learns a deformable 3D shape representing an object's category and captures its variations in different images. Due to the fitting-based inference procedure used by these methods, their 3D reconstruction is limited to categories, which lack detail and complex deformations. Therefore, Category-Specific Mesh Reconstruction, CMR \cite{cmr2018} proposes a prediction-based inference approach to 3D reconstruction, which predicts 3D shape, texture and camera pose from a collection of single-view images of deformable birds.

Other approaches with weaker supervision to 3D reconstruction learn a dense canonical surface mapping of objects from images with mask supervision and 3D template \cite{Kulkarni2019,Kulkarni2020}, however they do not learn to predict the 3D shape. In \cite{Xueting2020} the objects are represented as \textit{collection of deformable parts} and instead of keypoints on the canonical shape, the parts of canonical shape and their corresponding labels on the mask are applied.

The Unsupervised Category-Specific Mesh Reconstruction approach, U-CMR \cite{ucmr2020}, learns to predict shape, texture and pose using supervision in the form of object masks, and a category mean shape. We have noted that the U-CMR  \cite{ucmr2020} trained model does not perform optimally at inference. This could be because of the geometry concerning scale, translation, and rotation, is difficult for a neural network to learn.

To overcome this barrier, we designed a novel model with the same amount of supervision as U-CMR \cite{ucmr2020}, by replacing the camera pose prediction with prediction of a set of keypoints. These are then robustly matched with prototypical 3D keypoints to infer the pose. Our model applies standard neural layers and geometric optimization layers inside the same end-to-end framework inspired by recent works \cite{blind_pnp2008,ddn_2019,Campbell2020}. Just like \cite{ucmr2020}, we train deep neural networks with a collection of single views of a deformable category; birds, to predict 3D shapes, textures, and camera poses from single images, but we also extend the idea to work on video sequences by combining our method with LWL \cite{bhat2020learning}.

\vspace{-0.6em}
\subsection{Predicting 3D pose}
\vspace{-0.7em}

To obtain the pose of the camera capturing images, we need to predict relative scale, translation, and rotation where the choice of rotation representation has been studied in many recent research papers \cite{LevinsonECSKRM20,ZhouBLYL19,PeretroukhinGGR20,Bregier21}, the classic representations also include unit quaternion, Euler angles, and axis angle.

It is crucial to find a rotation representation, which works well with the neural networks. Neural networks, by design, generate feature vectors in the Euclidean space, and this limits their abilities to represent the desired output in many applications including 3D rotation matrix in a nonlinear and closed manifold for camera pose prediction  \cite{cmr2018,ucmr2020,MelekhovYKR17,EnLJ18}, objects \cite{XiangSNF18,MousavianAFK17} and humans \cite{KanazawaBJM18,Zhou0ZLW16} pose estimation.

A classical way to represent 3D rotation is unit quaternions \cite{cmr2018,ucmr2020},
which are popular due to their simplicity and lack of singularities. It has however, been suggested that their discontinuous nature (the double cover) is problematic \cite{ZhouBLYL19}. One proposed solution is to discretize the target space to solve a classification problem \cite{KanezakiMN18,KehlMTIN17}, however, this limits the precision of prediction as the number of classes required for accurate estimation could increase largely.

Alternatively \cite{LevinsonECSKRM20} proposed learning a 9D rotation representation that is orthogonalized using a Singular Value Decomposition (SVD). They showed that this 9D representation performs better than another proposed 6D representation, which is mapped onto SO(3) via a partial Gram-Schmidt procedure \cite{ZhouBLYL19}.

Another approach to estimating 3D rotation is to train an intermediate representation model in combination with the classical computer vision techniques to extract the geometric pose from a set of 2D-3D correspondences. In \cite{CrivellaroRVYFL15}, they used 2D projections of a preset controlled 3D points to solve a PnP problem for camera pose estimation.

Similar to \cite{CrivellaroRVYFL15}, we propose a model to predict camera poses from a set of 2D keypoints, but we apply an intermediate deep keypoint prediction network \cite{yu2018learning} to predict 2D keypoints. The 2D keypoints and their correspondences on the object’s 3D shape predicted by the mesh reconstruction network are then used by a robust PnP estimator \cite{Campbell2020} to extract the camera poses. In the experiments, we will compare our approach with \cite{ZhouBLYL19,LevinsonECSKRM20}, as well as a quaternion representation with a continuous loss, as used in \cite{cmr2018,ucmr2020}.

\subsection{Real-time 3D shape reconstruction from videos}
\vspace{-0.5em}

A recent research study \cite{Li2020OnlineAF} tried to adapt a pre-trained CMR \cite{cmr2018} for mesh reconstruction of non-rigid objects captured in naturalistic environments (e.g., animals) to illustrate a balance between model generalization and specialization. However, this approach is non-causal and the model requires a complete video sequence at inference time to be adapted. Moreover, in contrast to our online causal approach, which applies a prediction-based procedure at inference time, \cite{Li2020OnlineAF} uses a fitting-based procedure at inference time.

Our online approach to infer 3D object's shape, texture and pose from videos necessitates the detection of target object(s) from image frames. Detecting single or multiple objects in consecutive frames of a video sequence is quite challenging due to the varying scale and translation, and in the presence of occlusion. To tackle this problem, we utilize a tracker, Learning What to Learn (LWL) \cite{bhat2020learning} to track single and multiple birds of any kind in frames of YouTubeVos and Davis \cite{xu2018youtube} videos. Using LWL, height and width of a bounding box are predicted for each bird and the corresponding image is cropped to the bounding box, and the image patches are received as inputs by the reconstruction model.

\vspace{-0.7em}
\section{Baseline approach to shape and texture reconstruction}
\vspace{-0.8em}
\label{sec:ucmr}

This section presents the baseline approach to the shape and texture reconstruction framework U-CMR \cite{ucmr2020}, which receives an RGB image and an object mask as input, and learns a category-specific mean shape (e.g., a prototypical bird representing all birds) its texture, and per-frame deformation, and a camera pose distribution.

Running U-CMR, it first optimizes a randomly initialized \textit{camera-multiplex}, which is a representation of the distribution over cameras $\pi_k$, to best explain the image given the current shape and texture. Next, shape and texture models are trained applying (\ref{eq:1}), and the multiplex containing a set of best cameras pruned from the previous step is optimized using the camera update loss computed from the rendered masks, which is detached from the total loss (\ref{eq:1}). Finally, the best camera of the multiplex is used to train pose prediction network. Total loss of U-CMR is calculated as:  
\begin{equation}\label{eq:1}
	\mathcal{L}_{\textup{total}} = \sum_{k} p_k (\mathcal{L}_{\textup{mask},k} + \mathcal{L}_{\textup{pixel},k}) + \mathcal{L}_{\textup{def}} + \mathcal{L}_{\textup{lap}},
\end{equation}
where $\mathcal{L}_{\textup{mask},k}=\|S-\tilde{S}_k\|^2_2+dt(S)*\tilde{S}_k$ is the silhouette mask loss and $\mathcal{L}_{\textup{pixel},k}=\textup{dist}(\tilde{I}_k \odot S, I \odot S)$ is the image reconstruction loss computed from foreground regions. $S$ and $\tilde{S}_k$ are the ground-truth mask and the mask rendered from camera $\pi_k$, respectively. $dt(S)$ is the unidirectional distance transform of the ground-truth mask. $I$ and $\tilde{I}_k$ are the RGB image and the image rendered from camera $\pi_k$, respectively.

Moreover, in (\ref{eq:1}), $\mathcal{L}_{\textup{lap}}=\|V_i-\frac{1}{|\mathcal{N}(i)|} \sum_{j \in \mathcal{N}(i)}V_j\|^2$ is a graph-laplacian smoothness prior on the shape that penalizes the vertices $i$ that are far away from the centroid of their neighboring vertices $\mathcal{N}(i)$. For deformable objects like birds, it is beneficial to regularize the deformations to avoid arbitrary large deformations from the mean shape by adding the energy term $\mathcal{L}_{\textup{def}}=\|\Delta_V\|$. Finally, the probability of a camera $\pi_k$ be the optimal one is calculated by using $p_k=\frac{e^{-\frac{\mathcal{L}_k}{\sigma}}}{\sum_{j}e^{-\frac{\mathcal{L}_j}{\sigma}}}$, where $\mathcal{L}_k=\mathcal{L}_{\textup{mask},k} + \mathcal{L}_{\textup{pixel},k}$.

\vspace{-0.75em}
\section{Camera pose prediction}
\vspace{-0.75em}

In \cite{ucmr2020} pose is represented with an orthographic camera, which is parameterized by scale, translation, and rotation. Hence one parameter is used to represent the scale and two for translation, and four parameters are used to represent rotation using unit quaternions.
We instead predict a set of keypoints to represent the pose. These keypoints correspond to positions on the category-specific mean shape in 3D. Given such correspondences, a camera pose optimization algorithm like RANSAC \cite{fischler1981random}, can be used to find the camera pose, as we will describe below.

\subsection{Solving perspective-n-points by keypoint heatmaps}
\vspace{-0.7em}

In the U-CMR framework, the object shapes are predicted as deformations of a model shape, which is common for all images. The model shape has a fixed set of vertices $\mathcal{V}$ and faces $\mathcal{F}$.

As a result, for all object instances, the vertices correspond to the same semantic locations in the object shapes. They only differ in their deformed position, and can therefore be regarded as 3D keypoints. By reducing the number of these keypoints using down-sampling, and learning them in the 2D images, we should be able to find the object's pose using classical approaches, e.g using robust PnP estimators. 

Assume we have defined a set of semantic keypoints distributed over the 3D shape:
\begin{equation}\label{eq:2}
	X_{kp} = \{ x_{i_1},x_{i_2},x_{i_3} ..., x_{i_N} \},
\end{equation}
where $X_{kp}$ represents the set of 3D keypoints and $N$ is the total number of randomly selected keypoints, which are selected using Farthest Point Sampling \cite{qi2017pointnet++} as uniformly distributed all over the predicted shape. As an example, if the object is a bird, then the keypoints could be positioned on the head, beak, neck, wings, torso, and feet/talons. 

Estimation of the position and orientation of a camera relative to a scene using only 2D image points and 3D scene points in a trainable network requires solving the perspective-n-point problem \cite{Campbell2020}. 
To this end, we generate proxy ground-truth heatmaps by mapping the predicted 3D keypoints (\ref{eq:2}) to images using the optimized camera multiplex best poses.

To model the uncertainty of the keypoint positions we employ a Gaussian function centered at the keypoint image position. These {\it proxy ground-truth} heatmaps can then be used to supervise a CNN designed to predict keypoint heatmaps. Our keypoints prediction loss is formulated as: 
\begin{equation}\label{eq:3}
	\theta = {\arg\min}_{\theta} \sum_{i}^{N} S_w \| S_h -\mathscr{F}(x_i;\theta)\|^2,
\end{equation}
where $\mathscr{F}(x_i;\theta)$ is a keypoint prediction network, described in section \ref{sec:kppred} and $S=[S_h, S_w]$ is a self-supervision block, described in section \ref{sec:ssb} in details.

According to (\ref{eq:3}), we apply a weighted least square loss function summed over all $N$ keypoints based on a proxy ground-truth heatmaps $S_h$ and a predicted heatmap $\mathscr{F}(x_i;\theta)$ to optimize parameter $\theta$ of our prediction network $\mathscr{F}$. 

\begin{figure}
	\centering
	\includegraphics[width=0.7\textwidth]{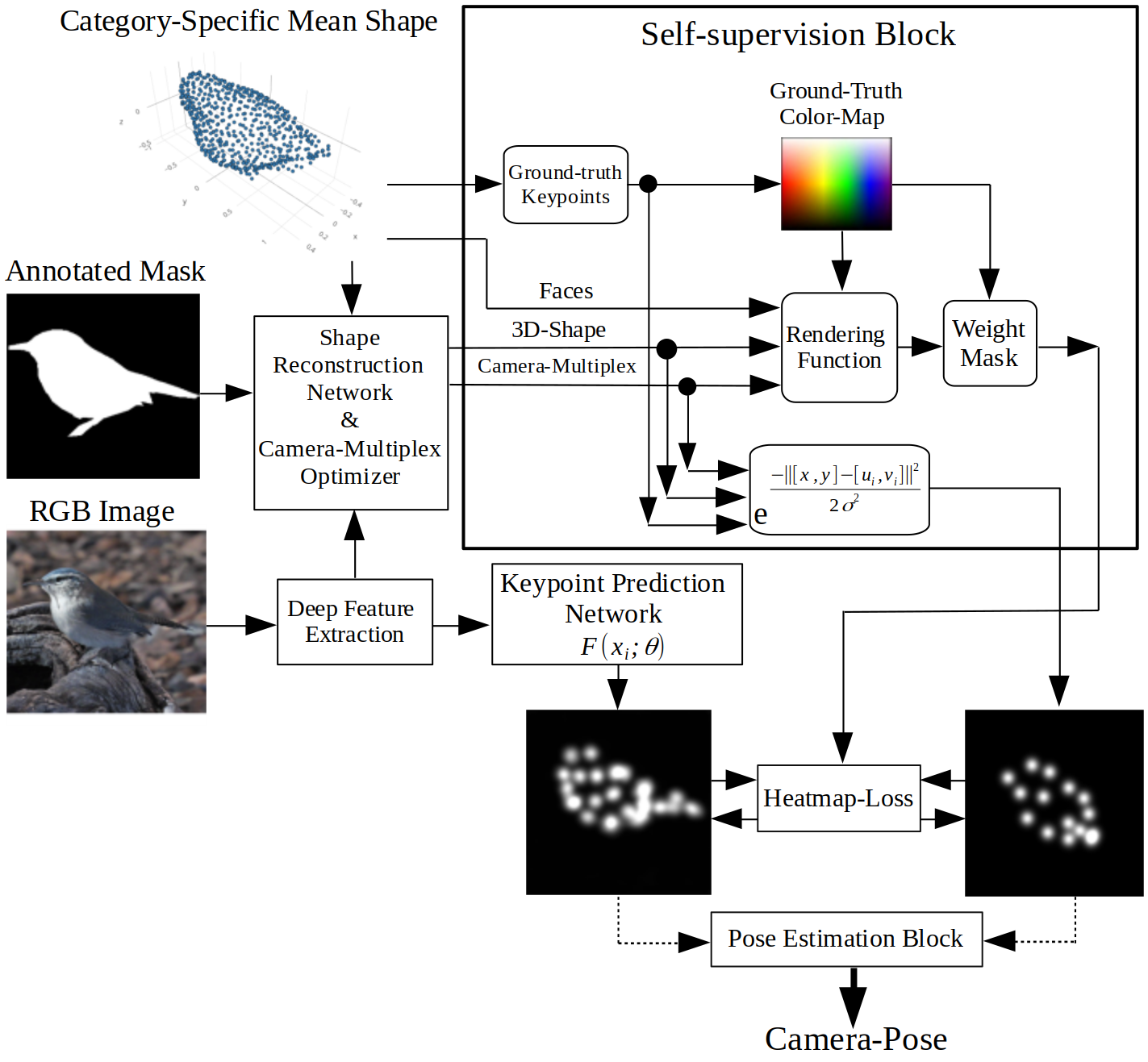}
	\caption{3D objects reconstruction framework using keypoints camera pose trainer. }
	\label{fig:arc_1}
\end{figure}
\subsubsection{Keypoint prediction network}
\vspace{-0.7em}
\label{sec:kppred}
The $\mathscr{F}(x_i;\theta)$ in (\ref{eq:3}) represents the deep keypoint prediction network \cite{yu2018learning}, which is  parametrized by $\theta$. It receives image deep features $x_i$ as input, and outputs a heatmap with one channel for each keypoint. We utilized a least squares loss function to train the parameters of $\mathscr{F}(x_i;\theta)$.

keypoint prediction network is composed of a Smooth network and a Border network. The Smooth network contains a channel attention block to select the discriminative features and a refinement residual block  to refine the feature map of each stage in the feature network. The Border network is applied to amplify the distinction of features using a semantic boundary to guide the feature learning. To enlarge the inter-class distinction of features the Border network learns a semantic boundary with an explicit supervision.

\vspace{-1.0em}
\subsubsection{Self-supervision block}
\vspace{-0.7em}
\label{sec:ssb}

We designed a self-supervision block $S$ to train our keypoints prediction network. This module is composed of two functional components $S_h$ and $S_w$ to generate a proxy ground-truth heatmaps and a weighting mask, respectively.

\paragraph{Proxy ground-truth heatmaps} To obtain the proxy ground-truth heatmaps, first we use $\hat{x_i} \in \mathbb{R}^3$, the 3D keypoints vertices randomly sampled over the object's predicted shape and $\pi$, the best camera pose reconstructed by the scale $s$, translation $t_{xy}$ and rotation $r_q$ from the shape reconstruction network, described in section \ref{sec:ucmr}, to project 3D keypoints $\hat{x_i} \in \mathbb{R}^3$ into their 2D correspondences $[u_i, v_i]$.

Next, we applied a Gaussian function to model the uncertainty of the locations of the 2D keypoints projections on the heatmap:      
\begin{equation}\label{eq:4}
	S_h = e^{\frac{-\|[x, y]-[u_i, v_i]\|^2}{2\sigma^2}}, 
\end{equation}
where $[x, y]$ are the 2D coordinates of the heatmap and $\sigma$ is a fixed value set to 0.05 in our experiments.

\paragraph{Weighted mask} To generate a weighted mask, the projected 2D vertices $[u_i, v_i]$ are then used to sample colors $\text{c}_\textup{sampled}$ from a rendered label texture $T_{l}$ (described in the following). The colors sampled from $T_{l}$ are compared with a ground-truth color map to generate a weighted mask $S_w$, which is a Boolean one-to-one correspondence matrix with at most one non-zero element in each row.

The $S_w$ outputs $1$ only if the predicted 3D keypoints $\hat{x_i} \in \mathbb{R}^3$, and their 2D correspondences on the proxy ground-truth heatmap $[u_i, v_i]$ are matching one-to-one based on a threshold applied to the difference value of their colors:
\begin{equation}\label{eq:5}
	S_w(\pi, \hat{x}_i, \text{C}_{\textup{map}}) = \delta_{\epsilon}[\|\text{c}_{\textup{sampled}}-\text{C}_{\textup{map}}\|<\epsilon],
\end{equation}
where $\delta_\epsilon[\textup{arg}]$ is an indicator, which returns $1$ if its argument is true, and zero otherwise. The colors in $\text{C}_{\textup{map}}$ are chosen to be more than $\epsilon$ apart, so (\ref{eq:5}) always works.

\paragraph{Rendered label texture} Due to the limitations in the used version of our renderer, $T_{l}$ is generated by considering a random set of $N$ keypoints distributed over the category specific mean shape shown in figure \ref{fig:arc_1}, a ground-truth color map $\text{C}_{\textup{map}}$ to assign a unique color to a particular keypoint and a ground-truth label texture $T_{\textup{gt}}$.  

Applying 3D mesh faces $F$ together with the predicted 3D mesh vertices $v$ and the optimized camera pose $\hat \pi$ from the shape reconstruction network (see section \ref{sec:ucmr}), the label texture is rendered and used to sample colors as $T_{l} = \text{R}(T_{\textup{gt}}, F, v, \hat \pi)$ applying a rendering function $\text{R}$.

\vspace{-1.0em}
\subsubsection{Pose estimation block}
\vspace{-0.7em} 
\label{sec:pnp}
When the keypoint prediction network is fully trained, we select those keypoints having the highest score from the predicted heatmaps together with their corresponding 3D points predicted by the shape reconstruction network and apply them to a robust PnP network \cite{Campbell2020} to solve the perspective-n-point problem to estimate the camera poses.

\vspace{-0.6em}
\section{Experiments}
\vspace{-0.5em}

To run experiments, first a multiplex containing cameras of varying poses is initialized and optimized for every image. Each camera is parameterized with 1D scale, 2D translation, and 3D angular rotation (azimuth, elevation, and cyclo-rotation). We randomly initialized 8 azimuth and 5 elevation scalar to create a multiplex of 40 cameras for each image.

In the second phase, first the multiplex of 40 cameras is shortly optimized using the camera update loss, which is computed from the rendered mask and is detached from the total loss (\ref{eq:1}), and then the multiplex is pruned to the best 4 cameras with the highest scores for every image. The shape and texture networks are trained during this phase applying the total loss (\ref{eq:1}), and the best camera of the multiplex.

Third is to train the camera pose prediction network, which predicts scale, translation, and rotation. We conducted our experiments to compare four different approaches to rotation representation. The first is to predict 4D unit quaternions by a CNN \cite{ucmr2020} . The second is 6D rotation representation mapped onto SO(3) via a partial Gram-Schmidt procedure \cite{ZhouBLYL19}, and the third is special orthogonalization using SVD \cite{LevinsonECSKRM20} based on 9D rotation representation. 
The fourth is our approach to camera pose prediction, which trains an intermediate keypoint prediction network, as described in section \ref{sec:kppred}.

\vspace{-0.4em}
\subsection{Off-line 3D object reconstruction from images}\label{sec:offline}
\vspace{-0.5em}

The first experiments are conducted by training single frames of the CUB \cite{wah2011caltech} image dataset applying different approaches to camera 3D pose prediction. 

\begin{figure*}
	\centering
	\begin{subfigure}{0.14\textwidth}
		\centering
		\includegraphics[width=\textwidth]{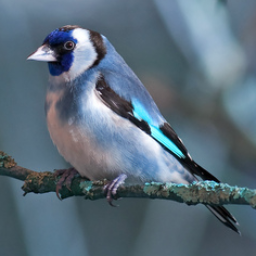}
		\label{fig:y equals x}
	\end{subfigure}
	\begin{subfigure}{0.14\textwidth}
		\centering
		\includegraphics[width=\textwidth]{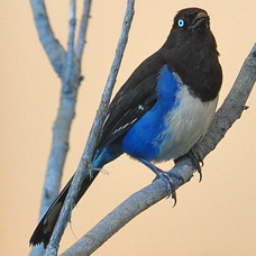}
		\label{fig:y equals x}
	\end{subfigure}
	\begin{subfigure}{0.15\textwidth}
		\centering
		\includegraphics[width=\textwidth]{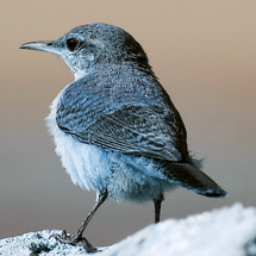}
		\label{fig:three sin x}
	\end{subfigure}
	\begin{subfigure}{0.14\textwidth}
		\centering
		\includegraphics[width=\textwidth]{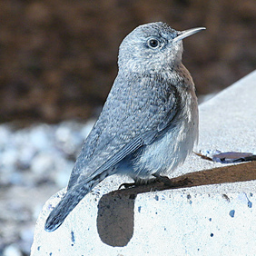}
		\label{fig:five over x}
	\end{subfigure}
	\vspace{-1em}
	\begin{subfigure}{0.14\textwidth}
		\centering
		\includegraphics[width=\textwidth]{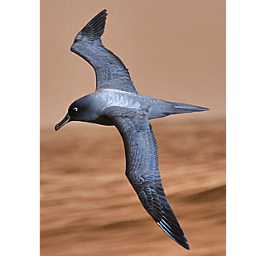}
		\label{fig:y equals x}
	\end{subfigure}
	\begin{subfigure}{0.14\textwidth}
		\centering
		\includegraphics[width=\textwidth]{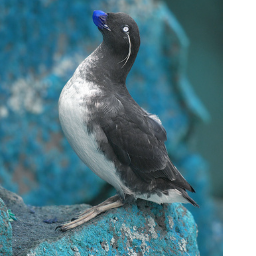}
		\label{fig:y equals x}
	\end{subfigure}
	\begin{subfigure}{0.14\textwidth}
		\centering
		\includegraphics[width=\textwidth]{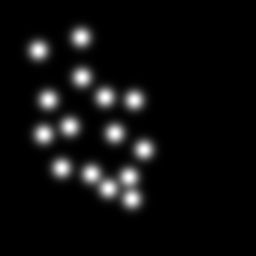}
		\label{fig:y equals x}
	\end{subfigure}
	\begin{subfigure}{0.14\textwidth}
		\centering
		\includegraphics[width=\textwidth]{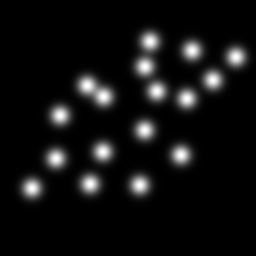}
		\label{fig:y equals x}
	\end{subfigure}
	\begin{subfigure}{0.14\textwidth}
		\centering
		\includegraphics[width=\textwidth]{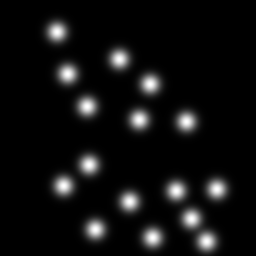}
		\label{fig:three sin x}
	\end{subfigure}
	\begin{subfigure}{0.14\textwidth}
		\centering
		\includegraphics[width=\textwidth]{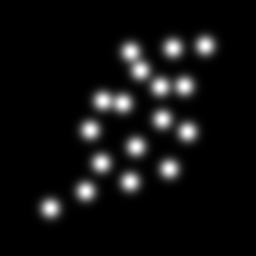}
		\label{fig:five over x}
	\end{subfigure}
	\vspace{-1em}
	\begin{subfigure}{0.14\textwidth}
		\centering
		\includegraphics[width=\textwidth]{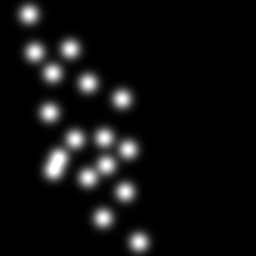}
		\label{fig:y equals x}
	\end{subfigure}
	\begin{subfigure}{0.14\textwidth}
		\centering
		\includegraphics[width=\textwidth]{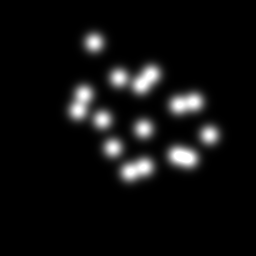}
		\label{fig:y equals x}
	\end{subfigure}
	\begin{subfigure}{0.14\textwidth}
		\centering
		\includegraphics[width=\textwidth]{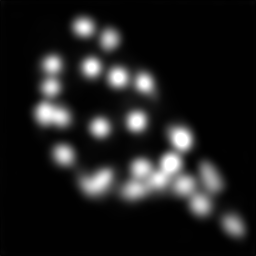}
		\label{fig:y equals x}
	\end{subfigure}
	\begin{subfigure}{0.14\textwidth}
		\centering
		\includegraphics[width=\textwidth]{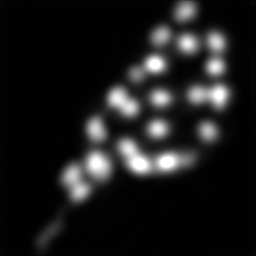}
		\label{fig:y equals x}
	\end{subfigure}
	\begin{subfigure}{0.14\textwidth}
		\centering
		\includegraphics[width=\textwidth]{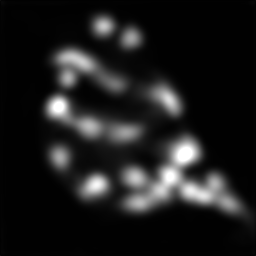}
		\label{fig:three sin x}
	\end{subfigure}
	\begin{subfigure}{0.14\textwidth}
		\centering
		\includegraphics[width=\textwidth]{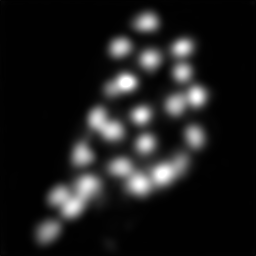}
		\label{fig:five over x}
	\end{subfigure}
	\vspace{-1em}
	\begin{subfigure}{0.14\textwidth}
		\centering
		\includegraphics[width=\textwidth]{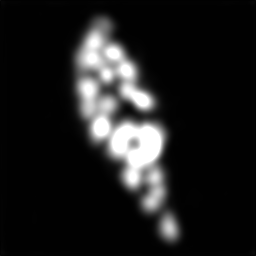}
		\label{fig:y equals x}
	\end{subfigure}
	\begin{subfigure}{0.14\textwidth}
		\centering
		\includegraphics[width=\textwidth]{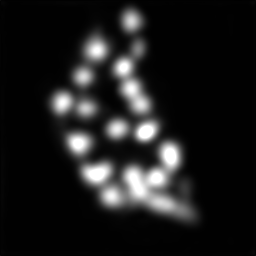}
		\label{fig:y equals x}
	\end{subfigure}
	\caption{First row shows original images of 6 different birds of CUB \cite{wah2011caltech} dataset. The second and third rows show the ground-truth heatmaps calculated by (\ref{eq:4}) and heatmaps predicted by the keypoints prediction network \cite{bhat2020learning}, respectively.}
	\label{fig:keypoints}
\end{figure*}

\begin{figure*}
	\centering
	\begin{subfigure}{0.14\textwidth}
		\centering
		\includegraphics[width=\textwidth]{European_Goldfinch_0037_33149_image.png}
		\label{fig:y equals x}
	\end{subfigure}
	\begin{subfigure}{0.14\textwidth}
		\centering
		\includegraphics[width=\textwidth]{Eastern_Towhee_0031_22233_image.png}
		\label{fig:y equals x}
	\end{subfigure}
	\begin{subfigure}{0.14\textwidth}
		\centering
		\includegraphics[width=\textwidth]{Rock_Wren_0065_188995_image.png}
		\label{fig:three sin x}
	\end{subfigure}
	\begin{subfigure}{0.14\textwidth}
		\centering
		\includegraphics[width=\textwidth]{Rock_Wren_0062_189045_image.png}
		\label{fig:five over x}
	\end{subfigure}
	\vspace{-1em}
	\begin{subfigure}{0.14\textwidth}
		\centering
		\includegraphics[width=\textwidth]{Sooty_Albatross_0049_796350_image.png}
		\label{fig:y equals x}
	\end{subfigure}
	\begin{subfigure}{0.14\textwidth}
		\centering
		\includegraphics[width=\textwidth]{Parakeet_Auklet_0064_795954_image.png}
		\label{fig:y equals x}
	\end{subfigure}
	\begin{subfigure}{0.14\textwidth}
		\centering
		\includegraphics[width=\textwidth]{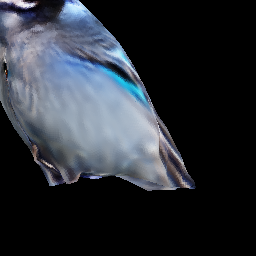}
		\label{fig:y equals x}
	\end{subfigure}
	\begin{subfigure}{0.14\textwidth}
		\centering
		\includegraphics[width=\textwidth]{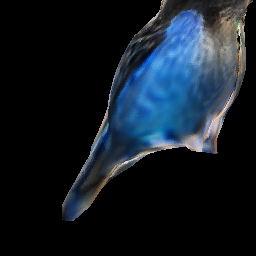}
		\label{fig:y equals x}
	\end{subfigure}
	\begin{subfigure}{0.14\textwidth}
		\centering
		\includegraphics[width=\textwidth]{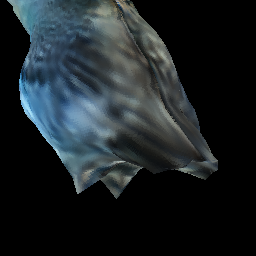}
		\label{fig:three sin x}
	\end{subfigure}
	\begin{subfigure}{0.14\textwidth}
		\centering
		\includegraphics[width=\textwidth]{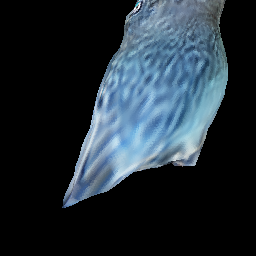}
		\label{fig:five over x}
	\end{subfigure}
	\vspace{-1em}
	\begin{subfigure}{0.14\textwidth}
		\centering
		\includegraphics[width=\textwidth]{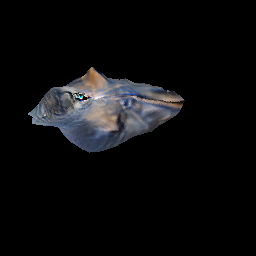}
		\label{fig:y equals x}
	\end{subfigure}
	\begin{subfigure}{0.14\textwidth}
		\centering
		\includegraphics[width=\textwidth]{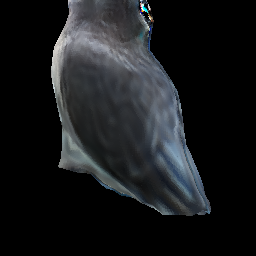}
		\label{fig:y equals x}
	\end{subfigure}
	\begin{subfigure}{0.14\textwidth}
		\centering
		\includegraphics[width=\textwidth]{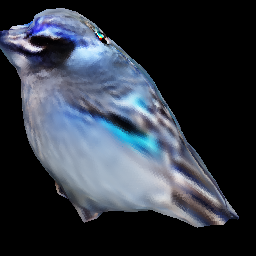}
		\label{fig:y equals x}
	\end{subfigure}
	\begin{subfigure}{0.14\textwidth}
		\centering
		\includegraphics[width=\textwidth]{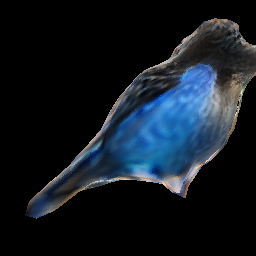}
		\label{fig:y equals x}
	\end{subfigure}
	\begin{subfigure}{0.14\textwidth}
		\centering
		\includegraphics[width=\textwidth]{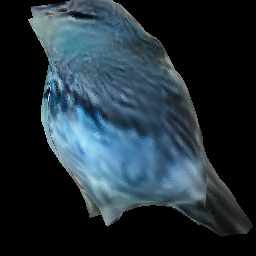}
		\label{fig:three sin x}
	\end{subfigure}
	\begin{subfigure}{0.14\textwidth}
		\centering
		\includegraphics[width=\textwidth]{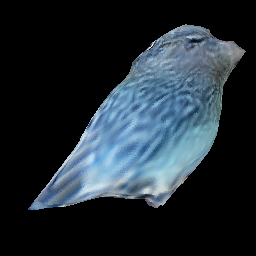}
		\label{fig:five over x}
	\end{subfigure}
	\vspace{-1em}
	\begin{subfigure}{0.14\textwidth}
		\centering
		\includegraphics[width=\textwidth]{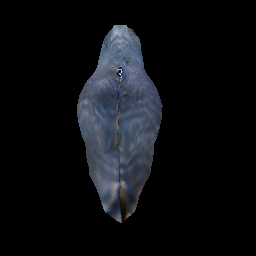}
		\label{fig:y equals x}
	\end{subfigure}
	\begin{subfigure}{0.14\textwidth}
		\centering
		\includegraphics[width=\textwidth]{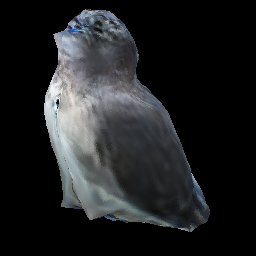}
		\label{fig:y equals x}
	\end{subfigure}
	\begin{subfigure}{0.14\textwidth}
		\centering
		\includegraphics[width=\textwidth]{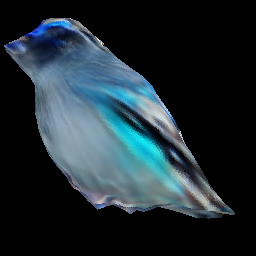}
		\label{fig:y equals x}
	\end{subfigure}
	\begin{subfigure}{0.14\textwidth}
		\centering
		\includegraphics[width=\textwidth]{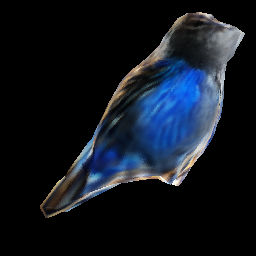}
		\label{fig:y equals x}
	\end{subfigure}
	\begin{subfigure}{0.14\textwidth}
		\centering
		\includegraphics[width=\textwidth]{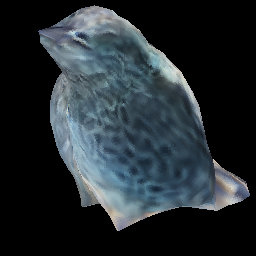}
		\label{fig:three sin x}
	\end{subfigure}
	\begin{subfigure}{0.14\textwidth}
		\centering
		\includegraphics[width=\textwidth]{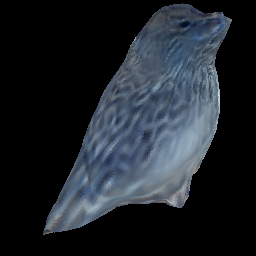}
		\label{fig:five over x}
	\end{subfigure}
	\vspace{-1em}
	\begin{subfigure}{0.14\textwidth}
		\centering
		\includegraphics[width=\textwidth]{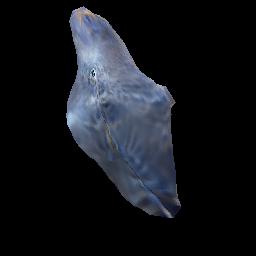}
		\label{fig:y equals x}
	\end{subfigure}
	\begin{subfigure}{0.14\textwidth}
		\centering
		\includegraphics[width=\textwidth]{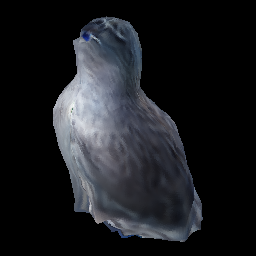}
		\label{fig:y equals x}
	\end{subfigure}
	\caption{Texture reconstruction. The first row shows original images of 6 different birds from the CUB \cite{wah2011caltech} dataset. Second row represents textures reconstructed using SfM camera poses for rendering. The third and fourth rows show the textures reconstructed when camera poses are predicted by unit quaternions and keypoints pose trainer, respectively.}
	\label{fig:textures}
\end{figure*}

The results of these experiments are shown in figures \ref{fig:keypoints} and figure \ref{fig:textures} and table \ref{tab:tab1}. According to figure \ref{fig:keypoints} the first row shows the original RGB images representing different types of birds with varying camera poses. The second and third rows show the ground-truth and predicted heatmaps, respectively. 

\begin{table}
	\centering
	\begin{tabular}{p{3cm}p{2cm}p{2cm}p{2cm}}
		Pose-Trainer & Mean-IoU & 3D-Angular-Error \\
		\hline 
		Quaternion (\cite{ucmr2020}) & 0.62 & $45.5^\circ$\\
		\hline 
		Gram-Schmidt \cite{ZhouBLYL19} & 0.55 & $60.4^\circ$\\
		\hline 
		SVD \cite{LevinsonECSKRM20} & 0.58 & $50.8^\circ$\\
		\hline
		Keypoint (proposed) & 0.70 & $40.5^\circ$\\
		\hline 
	\end{tabular}
	\caption{The results of a single 3D object reconstruction presented by mean intersection over union \textit{Mean-IoU} and \textit{3D-Angular-Error} calculated for the CUB \cite{wah2011caltech} test set when predicting camera pose by four different approaches, which are unit quaternions, Gram-Schmidt, special orthogonalization and the keypoint prediction. Note that the category-specific mesh reconstruction network CMR \cite{cmr2018} had a 3D-Angular error equal to $87.52^\circ$ when no viewpoint and keypoint supervision are used.}
	\label{tab:tab1}	
\end{table}

%

Mean intersection over union and 3D angular error in table \ref{tab:tab1}, also show the improved performance when compared to other methods. We achieved around $5^\circ$ reduction of 3D angular error, and an increase of mean intersection over union by almost $10\%$, compared to U-CMR \cite{ucmr2020}.

\begin{figure*}
	\centering
	\begin{subfigure}{0.10\textwidth}
		\centering
		\includegraphics[width=\textwidth]{European_Goldfinch_0037_33149_image.png}
		\label{fig:y equals x}
	\end{subfigure}
	\begin{subfigure}{0.10\textwidth}
		\centering
		\includegraphics[width=\textwidth]{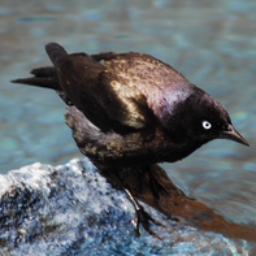}
		\label{fig:y equals x}
	\end{subfigure}
	\begin{subfigure}{0.10\textwidth}
		\centering
		\includegraphics[width=\textwidth]{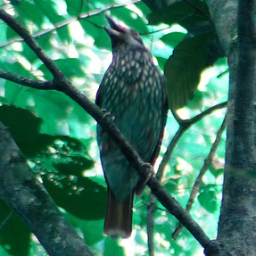}
		\label{fig:y equals x}
	\end{subfigure}
	\begin{subfigure}{0.10\textwidth}
		\centering
		\includegraphics[width=\textwidth]{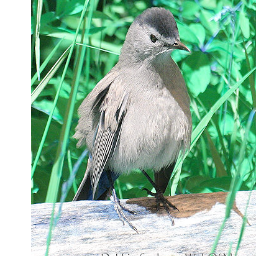}
		\label{fig:y equals x}
	\end{subfigure}
	\begin{subfigure}{0.10\textwidth}
		\centering
		\includegraphics[width=\textwidth]{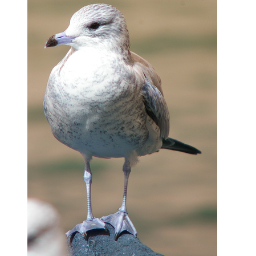}
		\label{fig:y equals x}
	\end{subfigure}
	\begin{subfigure}{0.10\textwidth}
		\centering
		\includegraphics[width=\textwidth]{Rock_Wren_0065_188995_image.png}
		\label{fig:three sin x}
	\end{subfigure}
	\begin{subfigure}{0.10\textwidth}
		\centering
		\includegraphics[width=\textwidth]{Rock_Wren_0062_189045_image.png}
		\label{fig:five over x}
	\end{subfigure}
	\vspace{-1em}
	\begin{subfigure}{0.10\textwidth}
		\centering
		\includegraphics[width=\textwidth]{Sooty_Albatross_0049_796350_image.png}
		\label{fig:y equals x}
	\end{subfigure}
	\begin{subfigure}{0.10\textwidth}
		\centering
		\includegraphics[width=\textwidth]{Parakeet_Auklet_0064_795954_image.png}
		\label{fig:y equals x}
	\end{subfigure}
	\begin{subfigure}{0.10\textwidth}
		\centering
		\includegraphics[width=\textwidth]{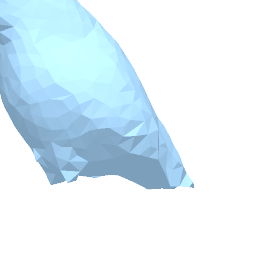}
		\label{fig:y equals x}
	\end{subfigure}
	\begin{subfigure}{0.10\textwidth}
		\centering
		\includegraphics[width=\textwidth]{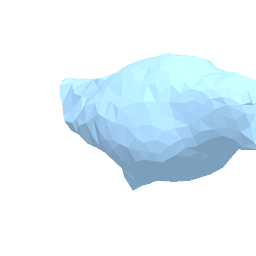}
		\label{fig:y equals x}
	\end{subfigure}
	\begin{subfigure}{0.10\textwidth}
		\centering
		\includegraphics[width=\textwidth]{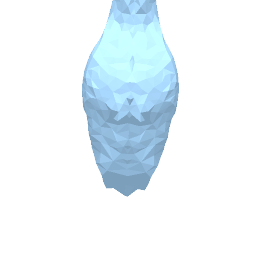}
		\label{fig:y equals x}
	\end{subfigure}
	\begin{subfigure}{0.10\textwidth}
		\centering
		\includegraphics[width=\textwidth]{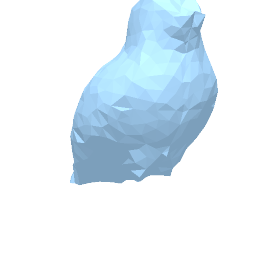}
		\label{fig:y equals x}
	\end{subfigure}
	\begin{subfigure}{0.10\textwidth}
		\centering
		\includegraphics[width=\textwidth]{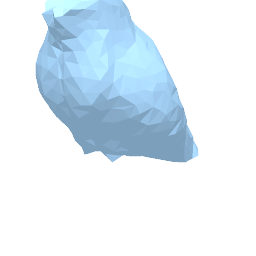}
		\label{fig:y equals x}
	\end{subfigure}
	\begin{subfigure}{0.10\textwidth}
		\centering
		\includegraphics[width=\textwidth]{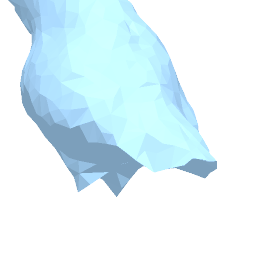}
		\label{fig:three sin x}
	\end{subfigure}
	\begin{subfigure}{0.10\textwidth}
		\centering
		\includegraphics[width=\textwidth]{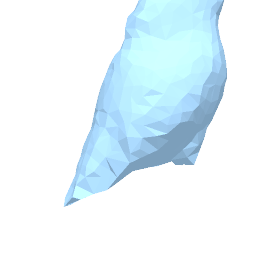}
		\label{fig:five over x}
	\end{subfigure}
	\vspace{-1em}
	\begin{subfigure}{0.10\textwidth}
		\centering
		\includegraphics[width=\textwidth]{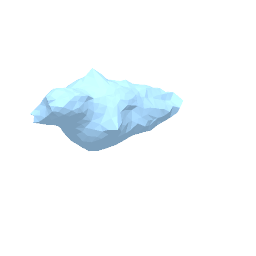}
		\label{fig:y equals x}
	\end{subfigure}
	\begin{subfigure}{0.10\textwidth}
		\centering
		\includegraphics[width=\textwidth]{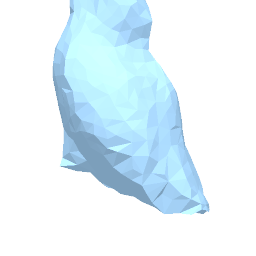}
		\label{fig:y equals x}
	\end{subfigure}
	\begin{subfigure}{0.10\textwidth}
		\centering
		\includegraphics[width=\textwidth]{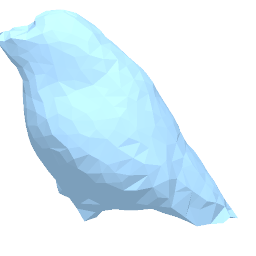}
		\label{fig:y equals x}
	\end{subfigure}
	\begin{subfigure}{0.10\textwidth}
		\centering
		\includegraphics[width=\textwidth]{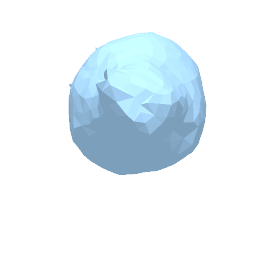}
		\label{fig:y equals x}
	\end{subfigure}
	\begin{subfigure}{0.10\textwidth}
		\centering
		\includegraphics[width=\textwidth]{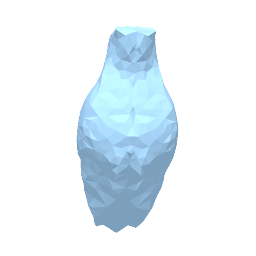}
		\label{fig:y equals x}
	\end{subfigure}
	\begin{subfigure}{0.10\textwidth}
		\centering
		\includegraphics[width=\textwidth]{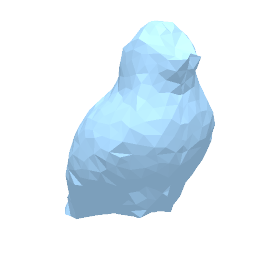}
		\label{fig:y equals x}
	\end{subfigure}
	\begin{subfigure}{0.10\textwidth}
		\centering
		\includegraphics[width=\textwidth]{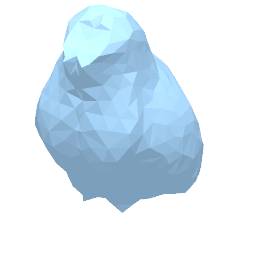}
		\label{fig:y equals x}
	\end{subfigure}
	\begin{subfigure}{0.10\textwidth}
		\centering
		\includegraphics[width=\textwidth]{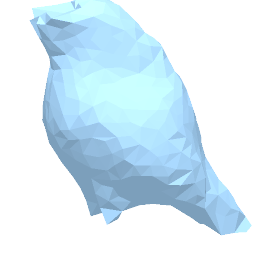}
		\label{fig:three sin x}
	\end{subfigure}
	\begin{subfigure}{0.10\textwidth}
		\centering
		\includegraphics[width=\textwidth]{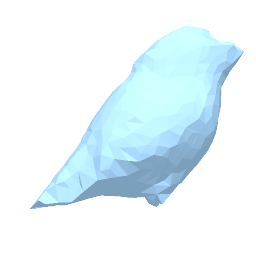}
		\label{fig:five over x}
	\end{subfigure}
	\vspace{-1em}
	\begin{subfigure}{0.10\textwidth}
		\centering
		\includegraphics[width=\textwidth]{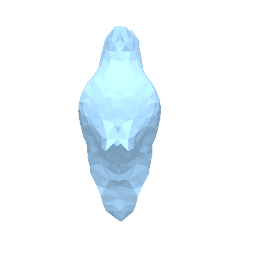}
		\label{fig:y equals x}
	\end{subfigure}
	\begin{subfigure}{0.10\textwidth}
		\centering
		\includegraphics[width=\textwidth]{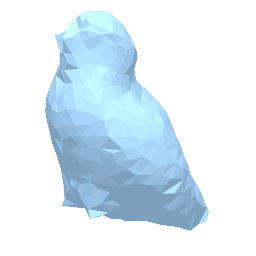}
		\label{fig:y equals x}
	\end{subfigure}
	\begin{subfigure}{0.10\textwidth}
		\centering
		\includegraphics[width=\textwidth]{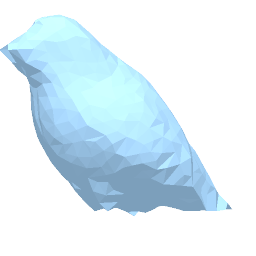}
		\label{fig:y equals x}
	\end{subfigure}
	\begin{subfigure}{0.10\textwidth}
		\centering
		\includegraphics[width=\textwidth]{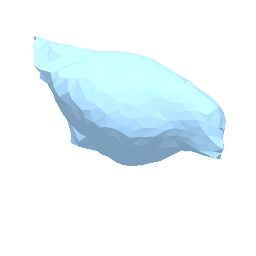}
		\label{fig:y equals x}
	\end{subfigure}
	\begin{subfigure}{0.10\textwidth}
		\centering
		\includegraphics[width=\textwidth]{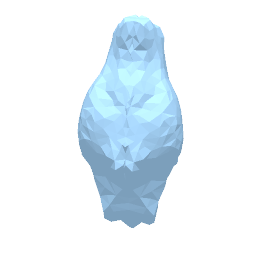}
		\label{fig:y equals x}
	\end{subfigure}
	\begin{subfigure}{0.10\textwidth}
		\centering
		\includegraphics[width=\textwidth]{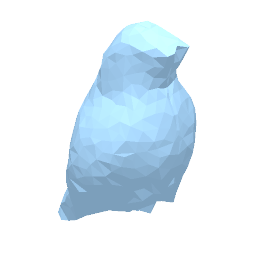}
		\label{fig:y equals x}
	\end{subfigure}
	\begin{subfigure}{0.10\textwidth}
		\centering
		\includegraphics[width=\textwidth]{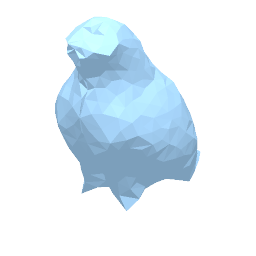}
		\label{fig:y equals x}
	\end{subfigure}
	\begin{subfigure}{0.10\textwidth}
		\centering
		\includegraphics[width=\textwidth]{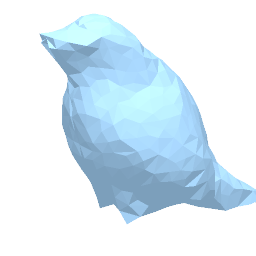}
		\label{fig:three sin x}
	\end{subfigure}
	\begin{subfigure}{0.10\textwidth}
		\centering
		\includegraphics[width=\textwidth]{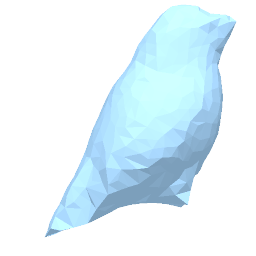}
		\label{fig:five over x}
	\end{subfigure}
	\vspace{-1em}
	\begin{subfigure}{0.10\textwidth}
		\centering
		\includegraphics[width=\textwidth]{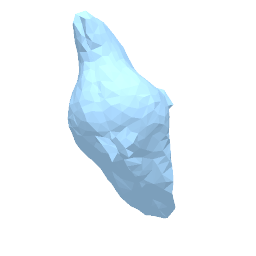}
		\label{fig:y equals x}
	\end{subfigure}
	\begin{subfigure}{0.10\textwidth}
		\centering
		\includegraphics[width=\textwidth]{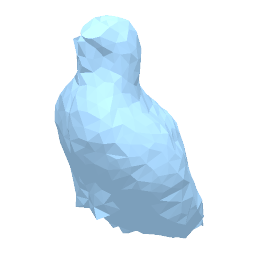}
		\label{fig:y equals x}
	\end{subfigure}
	\caption{3D mesh reconstruction. The first row shows original images of 9 different birds of CUB \cite{wah2011caltech} dataset, test set. The second row presents 3D meshes when applying the ground-truth camera pose SfM, available by dataset. Third row shows 3D shape when the camera poses are predicted using unit quaternions. The fourth row shows 3D shapes when the camera poses are predicted using keypoint correspondences.}
	\label{fig:3d_mesh}
\end{figure*}

In figure \ref{fig:3d_mesh}, qualitative examples of the mesh-reconstruction aided by our camera pose prediction, SfM poses, and quaternion based poses from U-CMR are shown on the CUB test images.

\begin{figure}
	\centering
	\includegraphics[width=\textwidth]{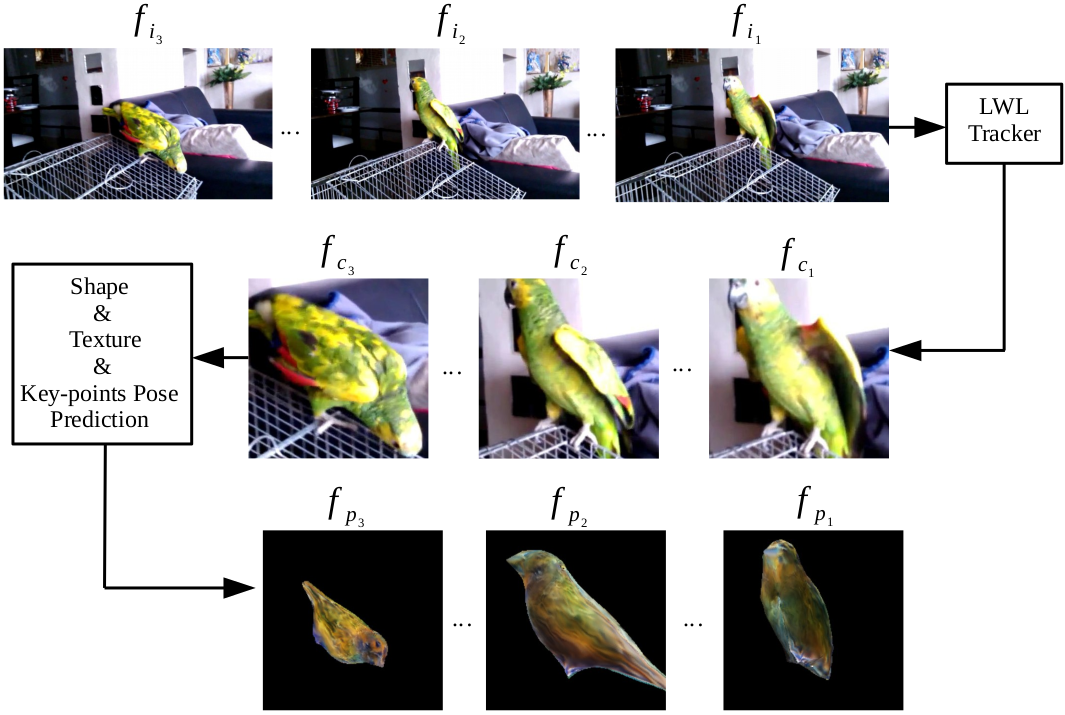}
	\caption{Online video object reconstruction framework. This example is from the YouTubeVos test set. The first row shows the original RGB images, and the second row shows the image patches generated by cropping using predicted bounding boxes from the LWL tracker. The third row shows the reconstructed shape and texture.}
	\label{fig:arc_2}
\end{figure}

\vspace{-1.0em}
\subsection{On-line inference 3D object reconstruction from videos} 
\vspace{-0.8em}

On-line experiments are conducted to infer 3D objects from video sequences containing single and multiple objects per image. Shown by figure \ref{fig:arc_2}, we used YouTubeVos and Davis \cite{xu2018youtube} datasets. Similar to the experiments presented in section \ref{sec:offline} four different approaches to camera pose prediction are applied here as well. We extracted bird category of YouTubeVos and Davis datasets to make it possible to compare our results with those presented by U-CMR \cite{ucmr2020}.

Inferring single and multiple objects from video sequences on-line is challenging due to varying position and orientation, and the occlusion. We use LWL \cite{bhat2020learning} to compute a bounding box from the predicted mask of the objects. Bounding box components are used to crop the frame and to create a patch as shown by figure \ref{fig:arc_2} second row. Image patches are used as the input of the reconstruction network, which predicts shape, texture, and camera pose shown by figure \ref{fig:arc_2} third row.

The objects' masks reconstructed by our approach to camera pose prediction and three other approaches are compared with the ground-truth masks and the results are presented in table \ref{tab:tab2}. Three metrics used for evaluation are the \textit{Jaccard-Mean}, which is the mean intersection over union, and the \textit{Jaccard-Recall}, which is the mean of fraction of values higher than a defined threshold, and the \textit{Jaccard-Decay}, which is the performance loss over time.

\begin{figure}
	\centering
	\includegraphics[width=0.8\textwidth]{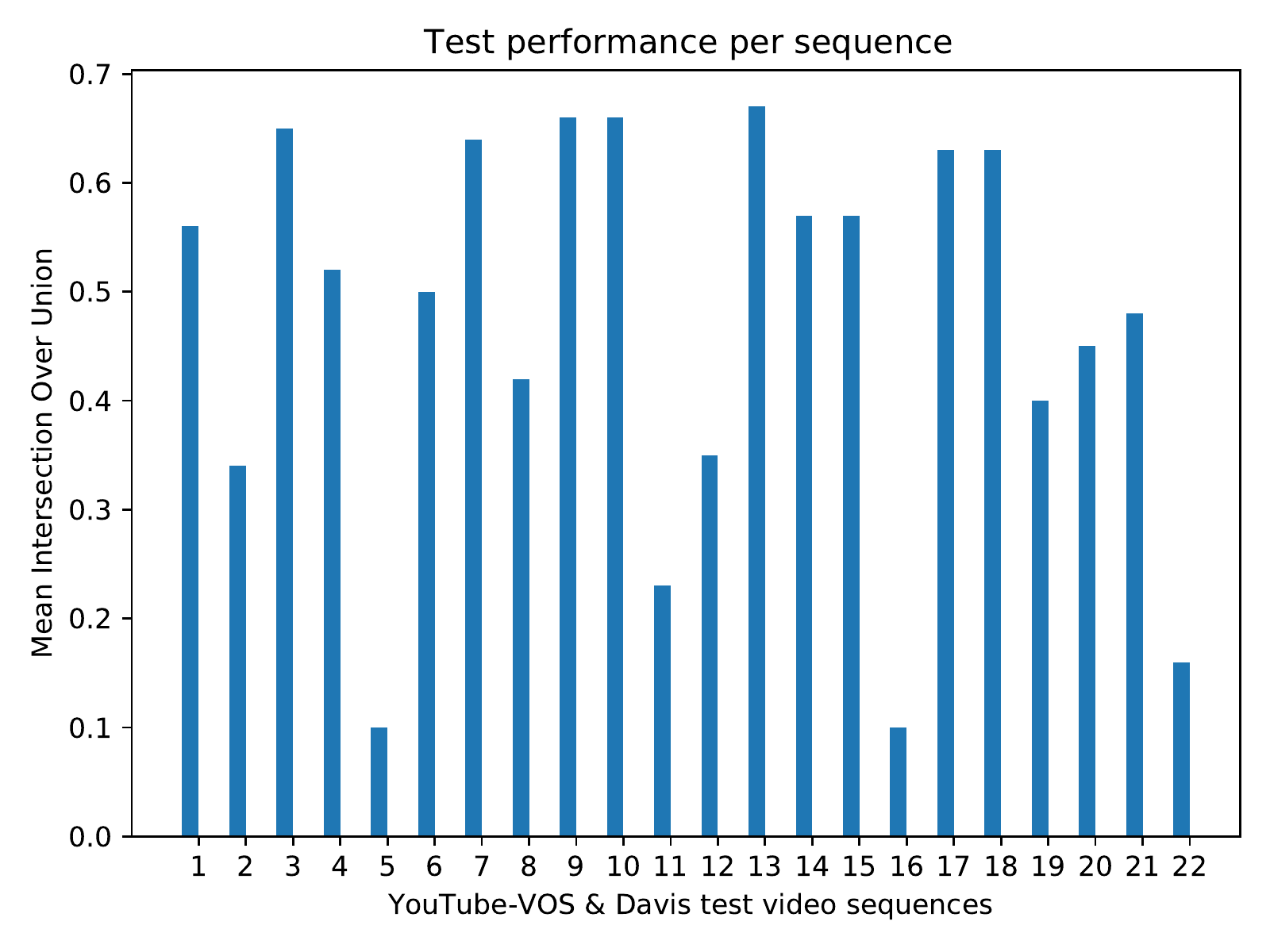}
	\caption{Mean intersection over union for 22 video sequences of YouTubeVos and Davis \cite{xu2018youtube} test sets.}
	\label{fig:bars}
\end{figure}

Figure \ref{fig:bars}, shows the results of on-line 3D shape prediction of 22 video sequences of YouTubeVos and Davis test sets. The results are calculated comparing ground-truth annotation with the model prediction and the mean intersection over union is computed by averaging the comparison matrix per visible object(s) per frame per sequence.



\vspace{-0.8em}
\section{Discussion}
\vspace{-0.8em}

We presented a novel approach to predict camera pose based on a self-supervised keypoint prediction network. We also constructed a simple on-line video object reconstruction system and demonstrated its performance on the task of on-line frame-by-frame prediction in previously unseen videos.

We trained our model on images of single birds with deformable shapes. To evaluate generalization abilities of our framework, we conducted experiments at inference using CUB \cite{wah2011caltech} test images, and YouTubeVos and Davis \cite{xu2018youtube} test videos. To distinguish birds from other object's categories, we propose using the LWL \cite{bhat2020learning} tracker trained on the YouTubeVos and Davis training set. This way the experiments are conducted on multiple objects per image.  

Compared to the online approach proposed in \cite{Li2020OnlineAF}, which is non-causal and thus requires a complete video sequence at inference applying a fitting-based procedure, our proposed model is causal, and applies a prediction-based procedure at inference.

Our model has never seen YouTubeVos and Davis videos before thus the overall performance is limited due to the minimum supervision and maximum generalization incentives. Moreover, it is trained on single frames of the CUB \cite{wah2011caltech} images without considering temporal constraints.  Therefore, we plan to improve the performance shown in table(\ref{tab:tab2}) and figure(\ref{fig:bars}) in future steps.

Our main objective in this study is to investigate the limitations of CNN when 3D predictions are required,
which is a major problem in objects’ 3D mesh prediction
and reconstruction. We also propose an approach to address these limitations and to improve the performance. \cite{sattler2019understanding} notes that direct pose regression is of limited use on rigid data, and that a classic local-feature+P3P-RANSAC works better. In contrast to this, our paper proposes an alternative network output, the keypoint heatmaps,
which are shown to be superior.

\vspace{-0.3em}
Our keypoint prediction network is self-supervised, and
the shape-reconstruction network has the same level of supervision as U-CMR \cite{ucmr2020}. In U-CMR this is called ”unsupervised”, but we agree that weak supervision is a better description.

U-CMR \cite{ucmr2020} predicts the geometric 3D pose by the neural networks since it represents rotation by the 4D unit quaternions. According to the results of U-CMR \cite{ucmr2020}, they got 3D angular error of $45^\circ$. The CMR \cite{cmr2018} also got 3D angular error of $87.52^\circ$ when no viewpoint and keypoint supervision were applied.

There is a significant opportunity in using CNNs to represent rotation, since they can effectively recognize patterns in the geometric data and thus reduce the search space and the influence of outliers. Therefore, their major applications are in classification tasks as they tend to produce clusters in the output space by applying dot product of the input vectors and the network's weights, which generate representations with peaks around the desired values.

\vspace{-0.3em}
On the contrary, the angular rotation requires a continuous representation space to cover $0^\circ$ to $360^\circ$, and if it is clustered into a definite number of classes to represent the rotation, the angular error increases as the precision of estimation decreases. Shown by the results of U-CMR \cite{ucmr2020}, 3D angular error of $45^\circ$ is consistent with 
the output representation space of rotation
being classified into 8 major clusters.   

\vspace{-0.3em}
To investigate other approaches to rotation representation, we applied the special orthogonalization by SVD \cite{LevinsonECSKRM20}, which claims to improve over Gram-Schmidt \cite{ZhouBLYL19} by a factor of two. We also observed the reduction of 3D angular error applying special orthogonalization by SVD \cite{LevinsonECSKRM20} compared to using the Gram-Schmidt \cite{ZhouBLYL19}. However, we didn't get any improvement using special orthogonalization by SVD compared to when the unit quaternions are used. A possible cause of this is the definition of the loss function for quaternions. If it is defined as $\min(d(q_p,q_r),d(q_p,-q_r))$, the double cover property of unit quaternions is properly taken into account, and the loss becomes continuous.

Our proposed model learns to predict the geometric pose from keypoints by self-supervised training of a deep keypoint prediction network. This way CNNs are used for feature extraction, which they are best designed for. Our results from table(\ref{tab:tab1}) show a significant improvement in pose prediction when compared with other related works. We gained mean intersection over the union of 0.70 and the 3D angular error of $40.50^\circ$, which is significantly superior when compared to predicting 4D unit quaternions \cite{ucmr2020}, special orthogonalization by SVD \cite{LevinsonECSKRM20} and Gram-Schmidt \cite{ZhouBLYL19}.

\begin{table}
	\centering
	\begin{tabular}{p{2.6cm}p{1.3cm}p{1.3cm}p{1.3cm}}
		Pose-Trainer  & Jaccard-Mean & Jaccard-Recall & Jaccard-Decay \\
		\hline
		Quaternion (\cite{ucmr2020}) &  0.45 & 0.48 & 0.020\\
		\hline 
		Gram-Schmidt\cite{ZhouBLYL19} & 0.40 & 0.27 & 0.007\\
		\hline 
		SVD\cite{LevinsonECSKRM20} & 0.44 & 0.43 & 0.028\\
		\hline 
		Keypoint\\ (proposed) & 0.45 & 0.53 & 0.018\\
		\hline 
	\end{tabular}
	\caption{Davis Jaccard mean, recall and decay for the YouTubeVos and Davis datasets. All birds are detected and tracked by LWL tracker and the corresponding image is cropped using the bounding box predicted by the tracker for each bird. The cropped patch is received by a pretrained mesh reconstruction model for camera pose prediction to solve perspective-n-point problem by employing keypoints correspondences.}	
	\label{tab:tab2}
\end{table}
\vspace{-1.0em}

\vspace{-0.0em}
\section{Conclusion}
\vspace{-0.5em}
In conclusion, this article proposes a framework to address camera pose prediction by self-supervised learning of keypoints correspondences, which can also reconstruct 3D objects from video sequences. Based on our results achieved by running experiments using both images and video sequences, our approach to camera pose prediction performs superior to those representing rotation with unit quaternions \cite{ucmr2020}, SVD orthogonalization \cite{LevinsonECSKRM20} and a partial Gram-Schmidt procedure \cite{ZhouBLYL19}. It is also interesting to note that unit quaternions with a continuous loss outperforms SVD and Gram-Schmidt.

The approach proposed to camera pose prediction presented in this paper learns to predict the camera poses self-supervised. This occurs by training a deep keypoint prediction network \cite{yu2018learning} using the 3D meshes predicted by a reconstruction model, and a pose optimized by a camera multiplex to generate the proxy ground-truth heatmaps.

Our architecture is capable of inferring 3D objects from video sequences despite the varying scale and translation and in the presence of occlusion. To this end, we applied the LWL tracker \cite{bhat2020learning}, which predicts a bounding box for the target objects(s) in every frame of a video sequence. 
This simple online video object reconstruction shows promise, and can be further improved by incorporating temporal consistency and online adaptation as in \cite{Li2020OnlineAF}.

\newpage
\bibliographystyle{splncs04}
\bibliography{mybibfile}

\newpage
\section{Appendix}
\begin{figure*}
	\centering
	\begin{subfigure}{0.14\textwidth}
		\centering
		\includegraphics[width=\textwidth]{European_Goldfinch_0037_33149_image.png}
		\label{fig:y equals x}
	\end{subfigure}
	\begin{subfigure}{0.14\textwidth}
		\centering
		\includegraphics[width=\textwidth]{Eastern_Towhee_0031_22233_image.png}
		\label{fig:y equals x}
	\end{subfigure}
	\begin{subfigure}{0.14\textwidth}
		\centering
		\includegraphics[width=\textwidth]{Rock_Wren_0065_188995_image.png}
		\label{fig:three sin x}
	\end{subfigure}
	\begin{subfigure}{0.14\textwidth}
		\centering
		\includegraphics[width=\textwidth]{Rock_Wren_0062_189045_image.png}
		\label{fig:five over x}
	\end{subfigure}
	\vspace{-1em}
	\begin{subfigure}{0.14\textwidth}
		\centering
		\includegraphics[width=\textwidth]{Sooty_Albatross_0049_796350_image.png}
		\label{fig:y equals x}
	\end{subfigure}
	\begin{subfigure}{0.14\textwidth}
		\centering
		\includegraphics[width=\textwidth]{Parakeet_Auklet_0064_795954_image.png}
		\label{fig:y equals x}
	\end{subfigure}
	\begin{subfigure}{0.14\textwidth}
		\centering
		\includegraphics[width=\textwidth]{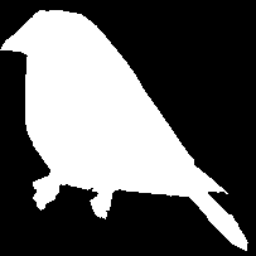}
		\label{fig:y equals x}
	\end{subfigure}
	\begin{subfigure}{0.14\textwidth}
		\centering
		\includegraphics[width=\textwidth]{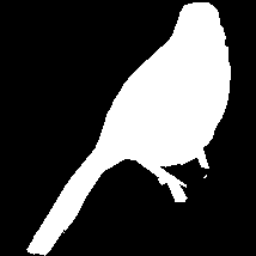}
		\label{fig:y equals x}
	\end{subfigure}
	\begin{subfigure}{0.14\textwidth}
		\centering
		\includegraphics[width=\textwidth]{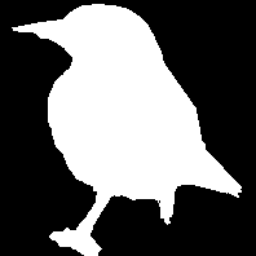}
		\label{fig:three sin x}
	\end{subfigure}
	\begin{subfigure}{0.14\textwidth}
		\centering
		\includegraphics[width=\textwidth]{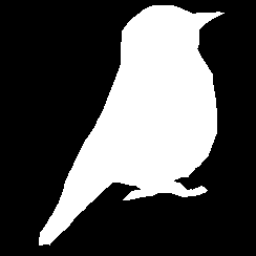}
		\label{fig:five over x}
	\end{subfigure}
	\vspace{-1em}
	\begin{subfigure}{0.14\textwidth}
		\centering
		\includegraphics[width=\textwidth]{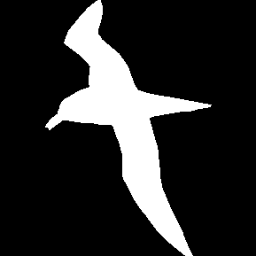}
		\label{fig:y equals x}
	\end{subfigure}
	\begin{subfigure}{0.14\textwidth}
		\centering
		\includegraphics[width=\textwidth]{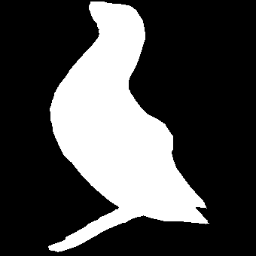}
		\label{fig:y equals x}
	\end{subfigure}
	\begin{subfigure}{0.14\textwidth}
		\centering
		\includegraphics[width=\textwidth]{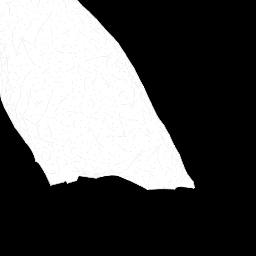}
		\label{fig:y equals x}
	\end{subfigure}
	\begin{subfigure}{0.14\textwidth}
		\centering
		\includegraphics[width=\textwidth]{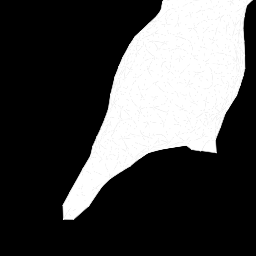}
		\label{fig:y equals x}
	\end{subfigure}
	\begin{subfigure}{0.14\textwidth}
		\centering
		\includegraphics[width=\textwidth]{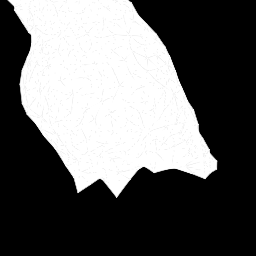}
		\label{fig:three sin x}
	\end{subfigure}
	\begin{subfigure}{0.14\textwidth}
		\centering
		\includegraphics[width=\textwidth]{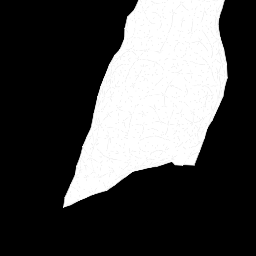}
		\label{fig:five over x}
	\end{subfigure}
	\vspace{-1em}
	\begin{subfigure}{0.14\textwidth}
		\centering
		\includegraphics[width=\textwidth]{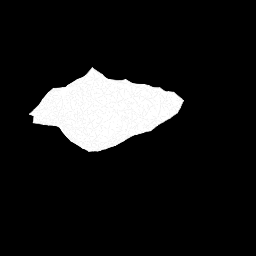}
		\label{fig:y equals x}
	\end{subfigure}
	\begin{subfigure}{0.14\textwidth}
		\centering
		\includegraphics[width=\textwidth]{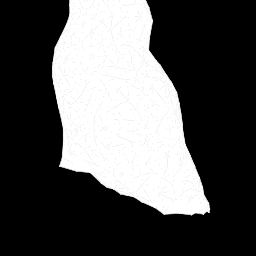}
		\label{fig:y equals x}
	\end{subfigure}
	\begin{subfigure}{0.14\textwidth}
		\centering
		\includegraphics[width=\textwidth]{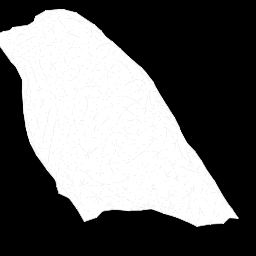}
		\label{fig:y equals x}
	\end{subfigure}
	\begin{subfigure}{0.14\textwidth}
		\centering
		\includegraphics[width=\textwidth]{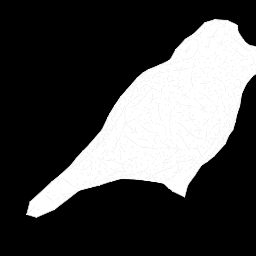}
		\label{fig:y equals x}
	\end{subfigure}
	\begin{subfigure}{0.14\textwidth}
		\centering
		\includegraphics[width=\textwidth]{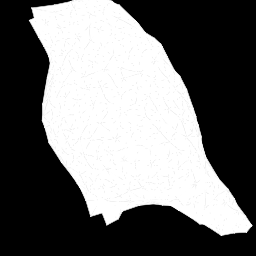}
		\label{fig:three sin x}
	\end{subfigure}
	\begin{subfigure}{0.14\textwidth}
		\centering
		\includegraphics[width=\textwidth]{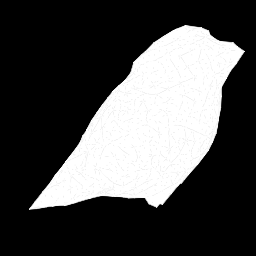}
		\label{fig:five over x}
	\end{subfigure}
	\vspace{-1em}
	\begin{subfigure}{0.14\textwidth}
		\centering
		\includegraphics[width=\textwidth]{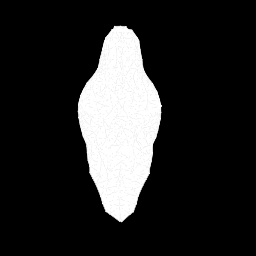}
		\label{fig:y equals x}
	\end{subfigure}
	\begin{subfigure}{0.14\textwidth}
		\centering
		\includegraphics[width=\textwidth]{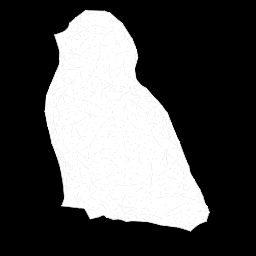}
		\label{fig:y equals x}
	\end{subfigure}
	\begin{subfigure}{0.14\textwidth}
		\centering
		\includegraphics[width=\textwidth]{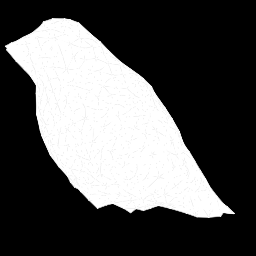}
		\label{fig:y equals x}
	\end{subfigure}
	\begin{subfigure}{0.14\textwidth}
		\centering
		\includegraphics[width=\textwidth]{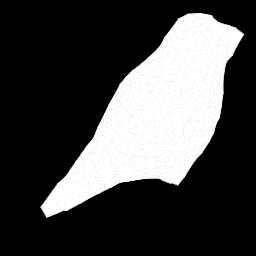}
		\label{fig:y equals x}
	\end{subfigure}
	\begin{subfigure}{0.14\textwidth}
		\centering
		\includegraphics[width=\textwidth]{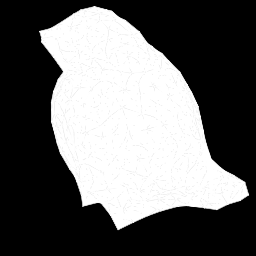}
		\label{fig:three sin x}
	\end{subfigure}
	\begin{subfigure}{0.14\textwidth}
		\centering
		\includegraphics[width=\textwidth]{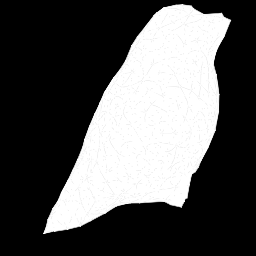}
		\label{fig:five over x}
	\end{subfigure}
	\vspace{-1em}
	\begin{subfigure}{0.14\textwidth}
		\centering
		\includegraphics[width=\textwidth]{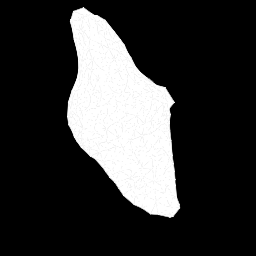}
		\label{fig:y equals x}
	\end{subfigure}
	\begin{subfigure}{0.14\textwidth}
		\centering
		\includegraphics[width=\textwidth]{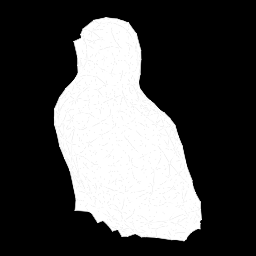}
		\label{fig:y equals x}
	\end{subfigure}
	\caption{Mask reconstruction. The first row shows original RBG images of 6 different birds of CUB \cite{wah2011caltech} dataset \cite{wah2011caltech} and the second row presents the ground-truth mask available in dataset. Third row represents rendered mask when SfM camera poses are used for rendering. The fourth and fifth rows show reconstructed masks when the camera poses predicted by using unit quaternions and keypoint correspondences are used, respectively.}
	\label{fig:masks}
\end{figure*}

\begin{figure*}
	\centering
	\begin{subfigure}{0.09\textwidth}
		\centering
		\includegraphics[width=\textwidth]{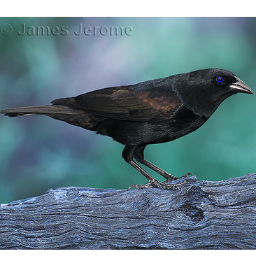}
		\label{fig:y equals x}
	\end{subfigure}
	\begin{subfigure}{0.09\textwidth}
		\centering
		\includegraphics[width=\textwidth]{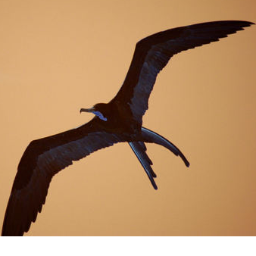}
		\label{fig:y equals x}
	\end{subfigure}
	\begin{subfigure}{0.09\textwidth}
		\centering
		\includegraphics[width=\textwidth]{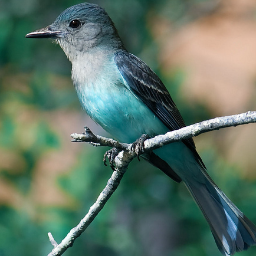}
		\label{fig:y equals x}
	\end{subfigure}
	\begin{subfigure}{0.09\textwidth}
		\centering
		\includegraphics[width=\textwidth]{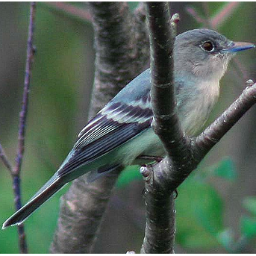}
		\label{fig:three sin x}
	\end{subfigure}
	\begin{subfigure}{0.09\textwidth}
		\centering
		\includegraphics[width=\textwidth]{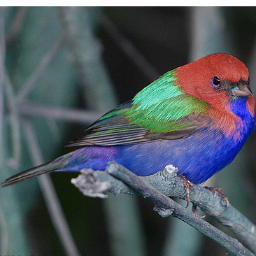}
		\label{fig:five over x}
	\end{subfigure}
	\vspace{-1em}
	\begin{subfigure}{0.09\textwidth}
		\centering
		\includegraphics[width=\textwidth]{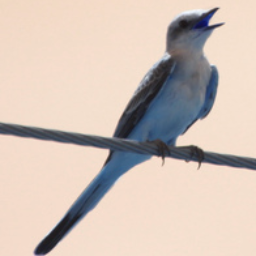}
		\label{fig:y equals x}
	\end{subfigure}
	\begin{subfigure}{0.09\textwidth}
		\centering
		\includegraphics[width=\textwidth]{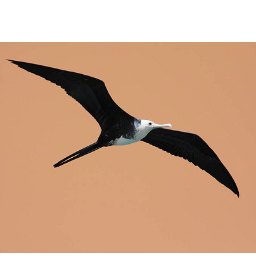}
		\label{fig:y equals x}
	\end{subfigure}
	\begin{subfigure}{0.09\textwidth}
		\centering
		\includegraphics[width=\textwidth]{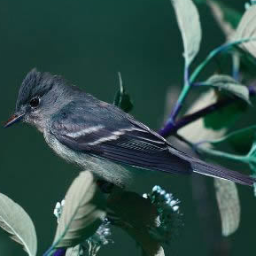}
		\label{fig:y equals x}
	\end{subfigure}
	\begin{subfigure}{0.09\textwidth}
		\centering
		\includegraphics[width=\textwidth]{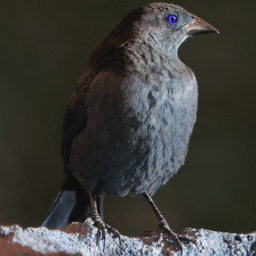}
		\label{fig:y equals x}
	\end{subfigure}
	\begin{subfigure}{0.09\textwidth}
		\centering
		\includegraphics[width=\textwidth]{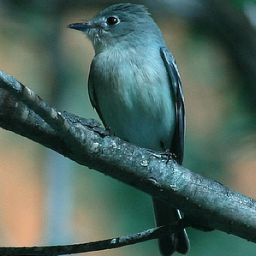}
		\label{fig:y equals x}
	\end{subfigure}
	\begin{subfigure}{0.09\textwidth}
		\centering
		\includegraphics[width=\textwidth]{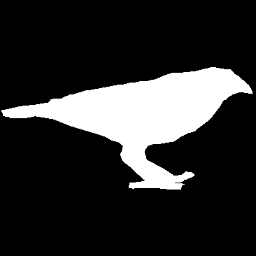}
		\label{fig:y equals x}
	\end{subfigure}
	\begin{subfigure}{0.09\textwidth}
		\centering
		\includegraphics[width=\textwidth]{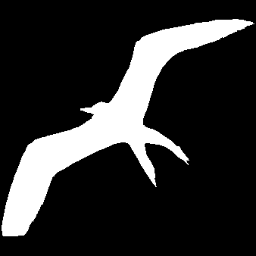}
		\label{fig:y equals x}
	\end{subfigure}
	\begin{subfigure}{0.09\textwidth}
		\centering
		\includegraphics[width=\textwidth]{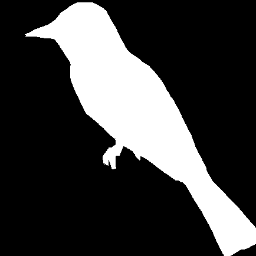}
		\label{fig:y equals x}
	\end{subfigure}
	\begin{subfigure}{0.09\textwidth}
		\centering
		\includegraphics[width=\textwidth]{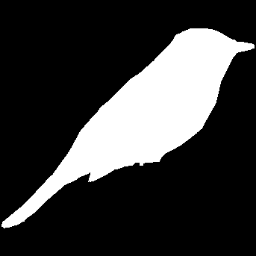}
		\label{fig:three sin x}
	\end{subfigure}
	\begin{subfigure}{0.09\textwidth}
		\centering
		\includegraphics[width=\textwidth]{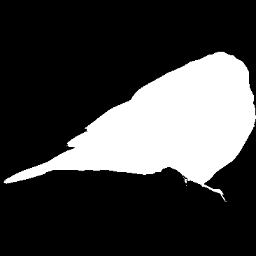}
		\label{fig:five over x}
	\end{subfigure}
	\vspace{-1em}
	\begin{subfigure}{0.09\textwidth}
		\centering
		\includegraphics[width=\textwidth]{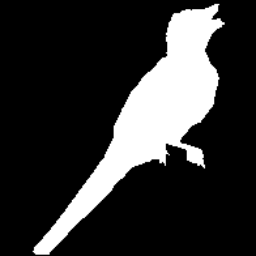}
		\label{fig:y equals x}
	\end{subfigure}
	\begin{subfigure}{0.09\textwidth}
		\centering
		\includegraphics[width=\textwidth]{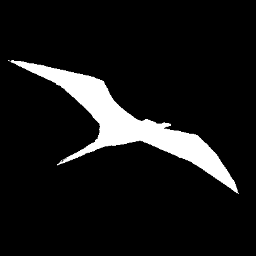}
		\label{fig:y equals x}
	\end{subfigure}
	\begin{subfigure}{0.09\textwidth}
		\centering
		\includegraphics[width=\textwidth]{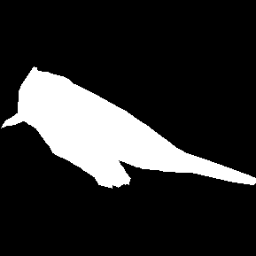}
		\label{fig:y equals x}
	\end{subfigure}
	\begin{subfigure}{0.09\textwidth}
		\centering
		\includegraphics[width=\textwidth]{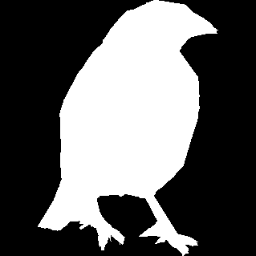}
		\label{fig:y equals x}
	\end{subfigure}
	\begin{subfigure}{0.09\textwidth}
		\centering
		\includegraphics[width=\textwidth]{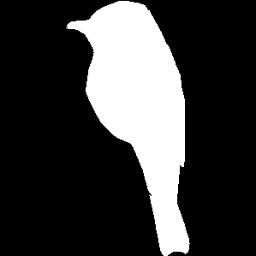}
		\label{fig:y equals x}
	\end{subfigure}
	\begin{subfigure}{0.09\textwidth}
		\centering
		\includegraphics[width=\textwidth]{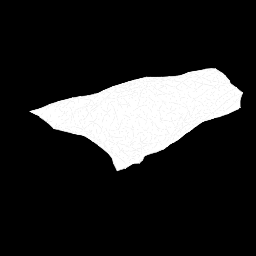}
		\label{fig:y equals x}
	\end{subfigure}
	\begin{subfigure}{0.09\textwidth}
		\centering
		\includegraphics[width=\textwidth]{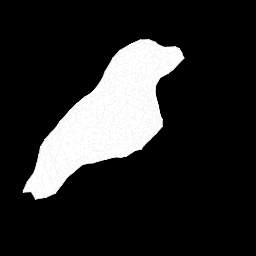}
		\label{fig:y equals x}
	\end{subfigure}
	\begin{subfigure}{0.09\textwidth}
		\centering
		\includegraphics[width=\textwidth]{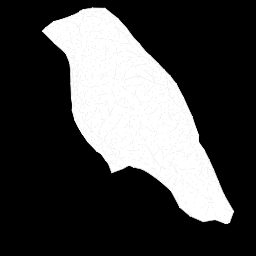}
		\label{fig:y equals x}
	\end{subfigure}
	\begin{subfigure}{0.09\textwidth}
		\centering
		\includegraphics[width=\textwidth]{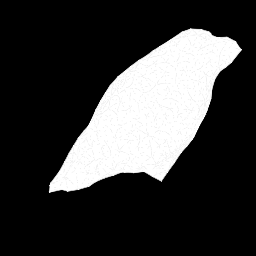}
		\label{fig:three sin x}
	\end{subfigure}
	\begin{subfigure}{0.09\textwidth}
		\centering
		\includegraphics[width=\textwidth]{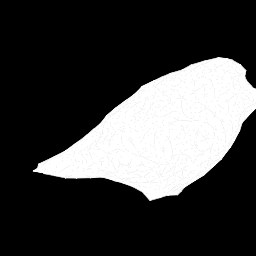}
		\label{fig:five over x}
	\end{subfigure}
	\vspace{-1em}
	\begin{subfigure}{0.09\textwidth}
		\centering
		\includegraphics[width=\textwidth]{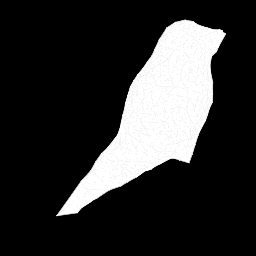}
		\label{fig:y equals x}
	\end{subfigure}
	\begin{subfigure}{0.09\textwidth}
		\centering
		\includegraphics[width=\textwidth]{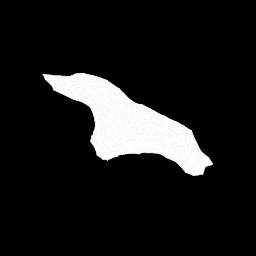}
		\label{fig:y equals x}
	\end{subfigure}
	\begin{subfigure}{0.09\textwidth}
		\centering
		\includegraphics[width=\textwidth]{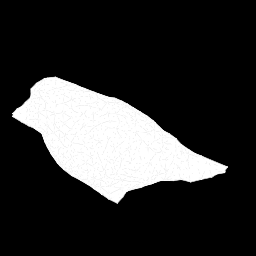}
		\label{fig:y equals x}
	\end{subfigure}
	\begin{subfigure}{0.09\textwidth}
		\centering
		\includegraphics[width=\textwidth]{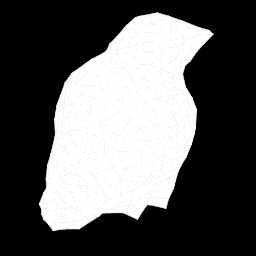}
		\label{fig:y equals x}
	\end{subfigure}
	\begin{subfigure}{0.09\textwidth}
		\centering
		\includegraphics[width=\textwidth]{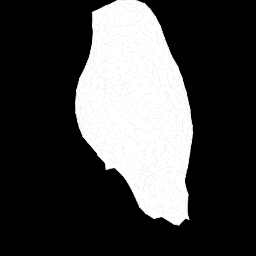}
		\label{fig:y equals x}
	\end{subfigure}
	\begin{subfigure}{0.09\textwidth}
		\centering
		\includegraphics[width=\textwidth]{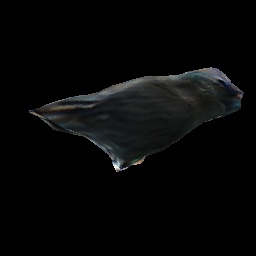}
		\label{fig:y equals x}
	\end{subfigure}
	\begin{subfigure}{0.09\textwidth}
		\centering
		\includegraphics[width=\textwidth]{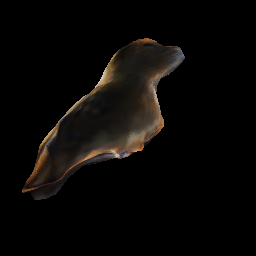}
		\label{fig:y equals x}
	\end{subfigure}
	\begin{subfigure}{0.09\textwidth}
		\centering
		\includegraphics[width=\textwidth]{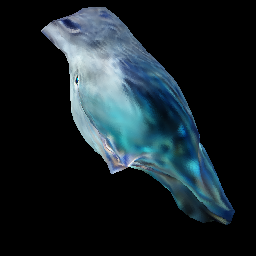}
		\label{fig:y equals x}
	\end{subfigure}
	\begin{subfigure}{0.09\textwidth}
		\centering
		\includegraphics[width=\textwidth]{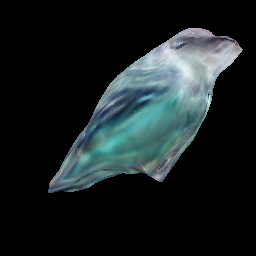}
		\label{fig:three sin x}
	\end{subfigure}
	\begin{subfigure}{0.09\textwidth}
		\centering
		\includegraphics[width=\textwidth]{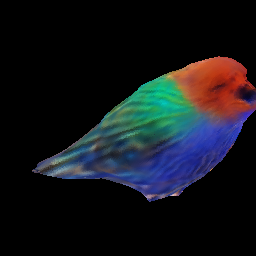}
		\label{fig:five over x}
	\end{subfigure}
	\vspace{-1em}
	\begin{subfigure}{0.09\textwidth}
		\centering
		\includegraphics[width=\textwidth]{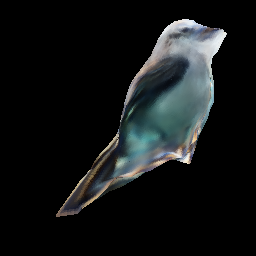}
		\label{fig:y equals x}
	\end{subfigure}
	\begin{subfigure}{0.09\textwidth}
		\centering
		\includegraphics[width=\textwidth]{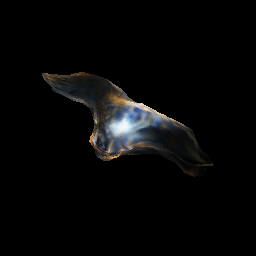}
		\label{fig:y equals x}
	\end{subfigure}
	\begin{subfigure}{0.09\textwidth}
		\centering
		\includegraphics[width=\textwidth]{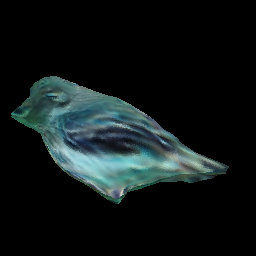}
		\label{fig:y equals x}
	\end{subfigure}
	\begin{subfigure}{0.09\textwidth}
		\centering
		\includegraphics[width=\textwidth]{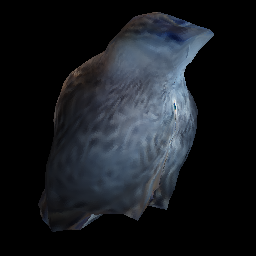}
		\label{fig:y equals x}
	\end{subfigure}
	\begin{subfigure}{0.09\textwidth}
		\centering
		\includegraphics[width=\textwidth]{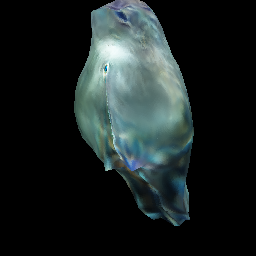}
		\label{fig:y equals x}
	\end{subfigure}
	\begin{subfigure}{0.09\textwidth}
		\centering
		\includegraphics[width=\textwidth]{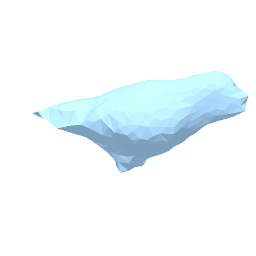}
		\label{fig:y equals x}
	\end{subfigure}
	\begin{subfigure}{0.09\textwidth}
		\centering
		\includegraphics[width=\textwidth]{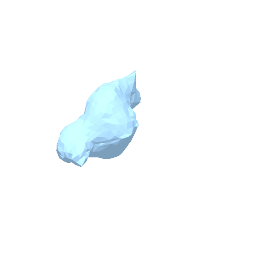}
		\label{fig:y equals x}
	\end{subfigure}
	\begin{subfigure}{0.09\textwidth}
		\centering
		\includegraphics[width=\textwidth]{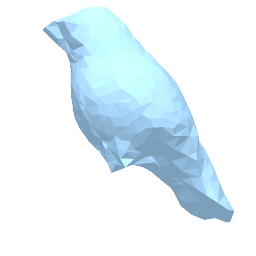}
		\label{fig:y equals x}
	\end{subfigure}
	\begin{subfigure}{0.09\textwidth}
		\centering
		\includegraphics[width=\textwidth]{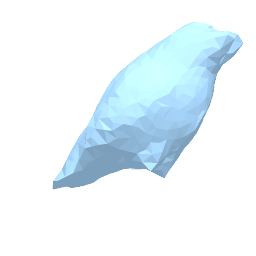}
		\label{fig:three sin x}
	\end{subfigure}
	\begin{subfigure}{0.09\textwidth}
		\centering
		\includegraphics[width=\textwidth]{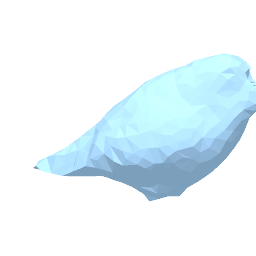}
		\label{fig:five over x}
	\end{subfigure}
	\vspace{-1em}
	\begin{subfigure}{0.09\textwidth}
		\centering
		\includegraphics[width=\textwidth]{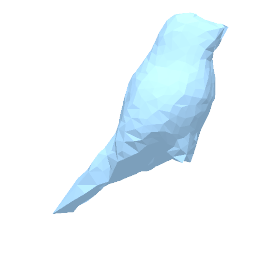}
		\label{fig:y equals x}
	\end{subfigure}
	\begin{subfigure}{0.09\textwidth}
		\centering
		\includegraphics[width=\textwidth]{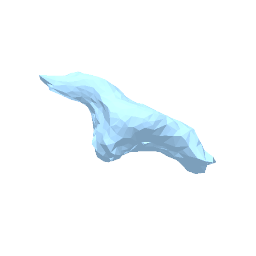}
		\label{fig:y equals x}
	\end{subfigure}
	\begin{subfigure}{0.09\textwidth}
		\centering
		\includegraphics[width=\textwidth]{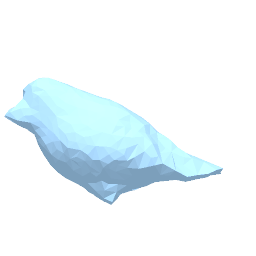}
		\label{fig:y equals x}
	\end{subfigure}
	\begin{subfigure}{0.09\textwidth}
		\centering
		\includegraphics[width=\textwidth]{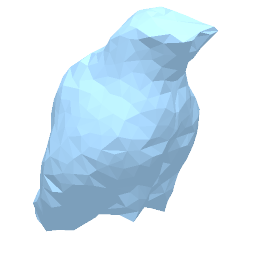}
		\label{fig:y equals x}
	\end{subfigure}
	\begin{subfigure}{0.09\textwidth}
		\centering
		\includegraphics[width=\textwidth]{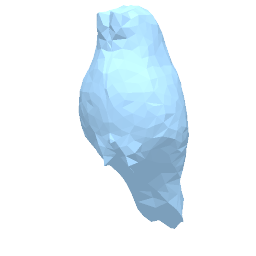}
		\label{fig:y equals x}
	\end{subfigure}
	\caption{Image reconstruction. The first row shows original images of 10 different birds from the CUB \cite{wah2011caltech} dataset. Second row represents ground-truth annotation available by dataset. The third and fourth rows show the masks and textures reconstructed when camera poses are predicted by keypoints, respectively. The fifth row represents the 3D shape reconstructed using camera pose predictions.}
	\label{fig:app_3}
\end{figure*}
%

\begin{figure*}
	\centering
	\begin{subfigure}{0.09\textwidth}
		\centering
		\includegraphics[width=\textwidth]{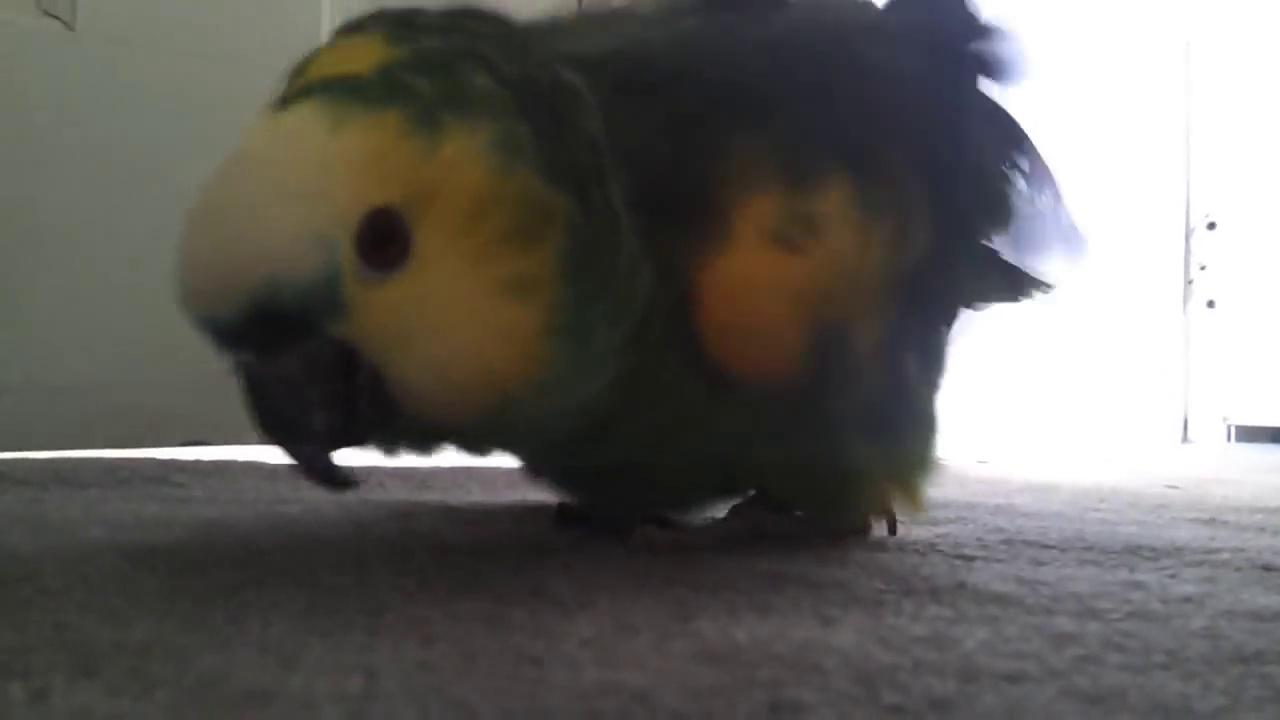}
		\label{fig:y equals x}
	\end{subfigure}
	\begin{subfigure}{0.09\textwidth}
		\centering
		\includegraphics[width=\textwidth]{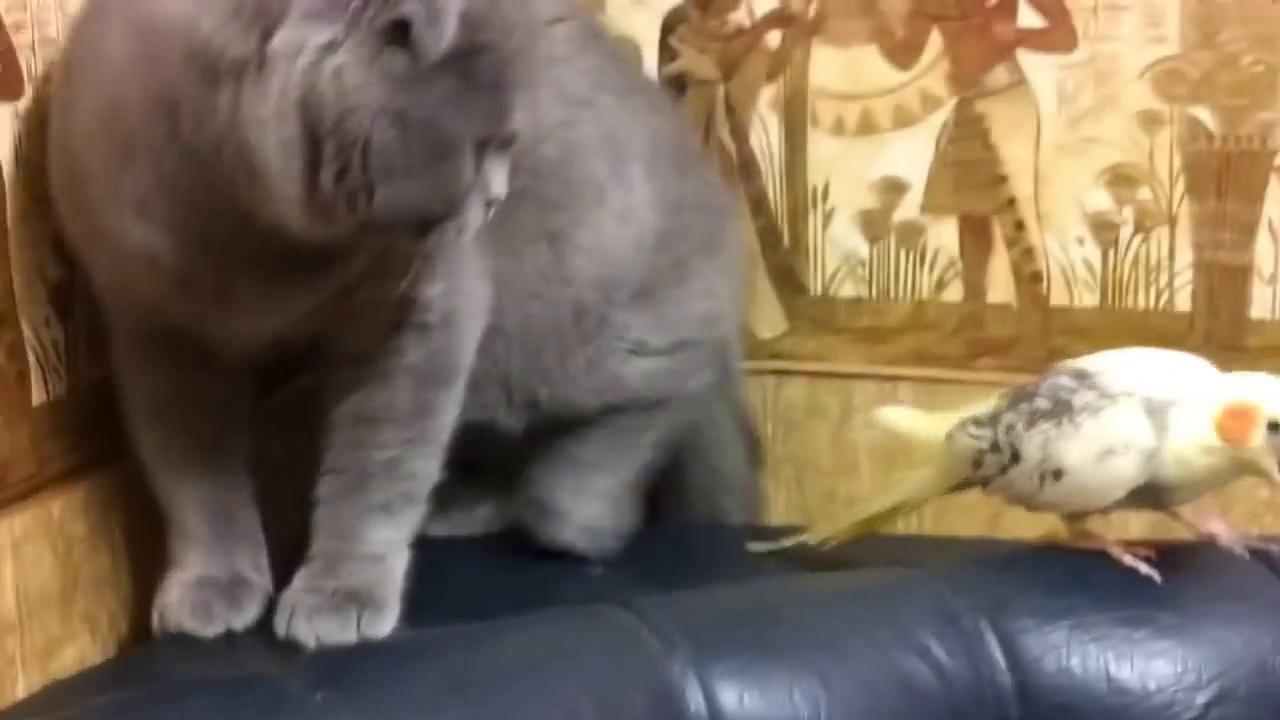}
		\label{fig:y equals x}
	\end{subfigure}
	\begin{subfigure}{0.09\textwidth}
		\centering
		\includegraphics[width=\textwidth]{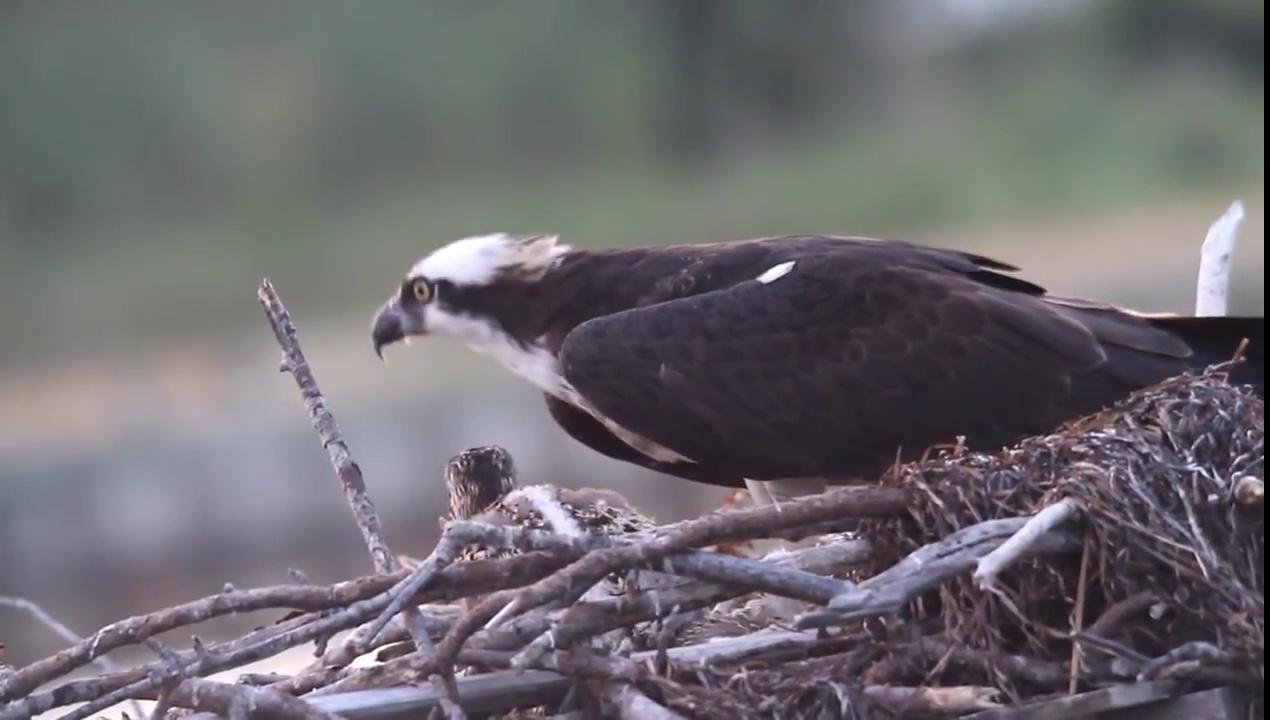}
		\label{fig:three sin x}
	\end{subfigure}
	\begin{subfigure}{0.09\textwidth}
		\centering
		\includegraphics[width=\textwidth]{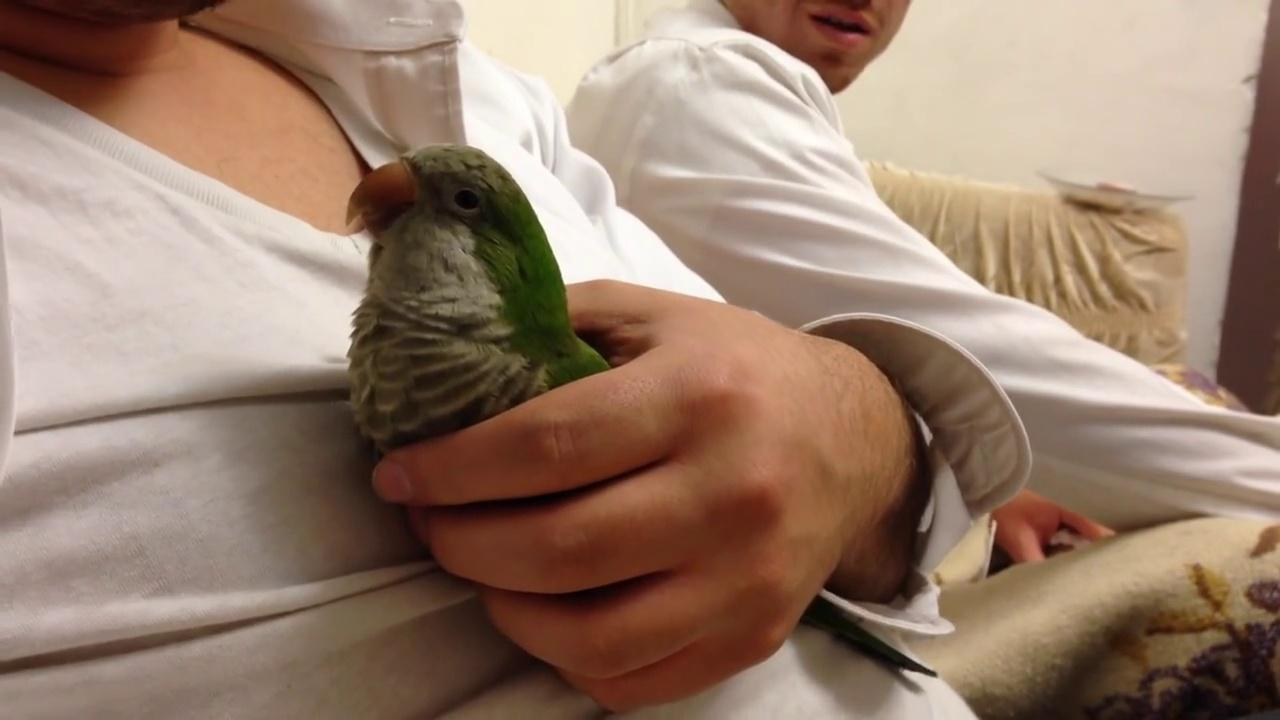}
		\label{fig:five over x}
	\end{subfigure}
	\vspace{-1em}
	\begin{subfigure}{0.09\textwidth}
		\centering
		\includegraphics[width=\textwidth]{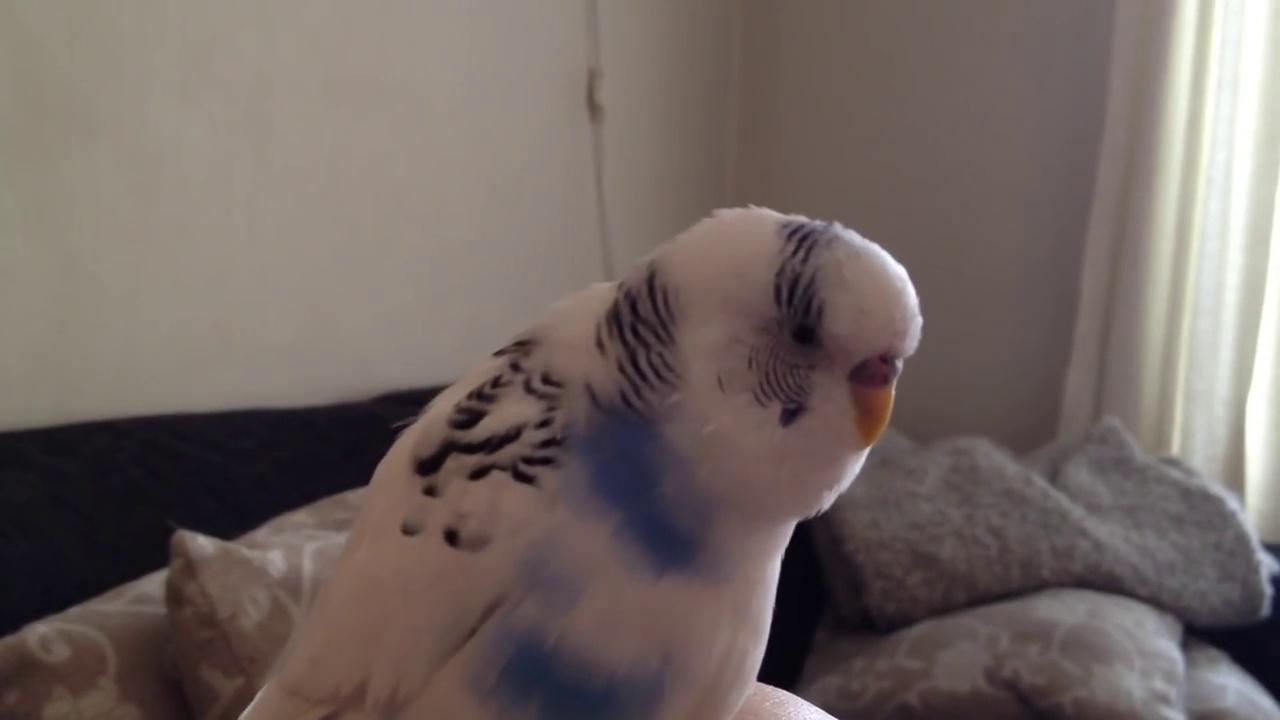}
		\label{fig:y equals x}
	\end{subfigure}
	\begin{subfigure}{0.09\textwidth}
		\centering
		\includegraphics[width=\textwidth]{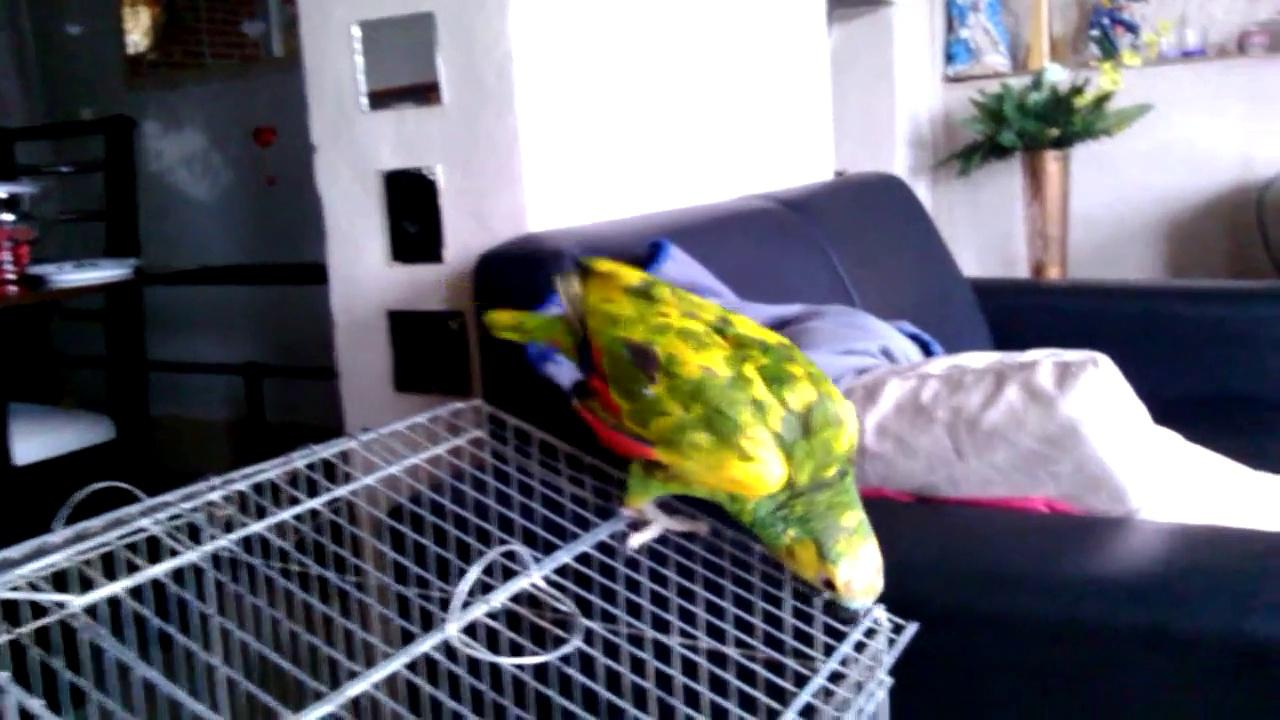}
		\label{fig:y equals x}
	\end{subfigure}
	\begin{subfigure}{0.09\textwidth}
		\centering
		\includegraphics[width=\textwidth]{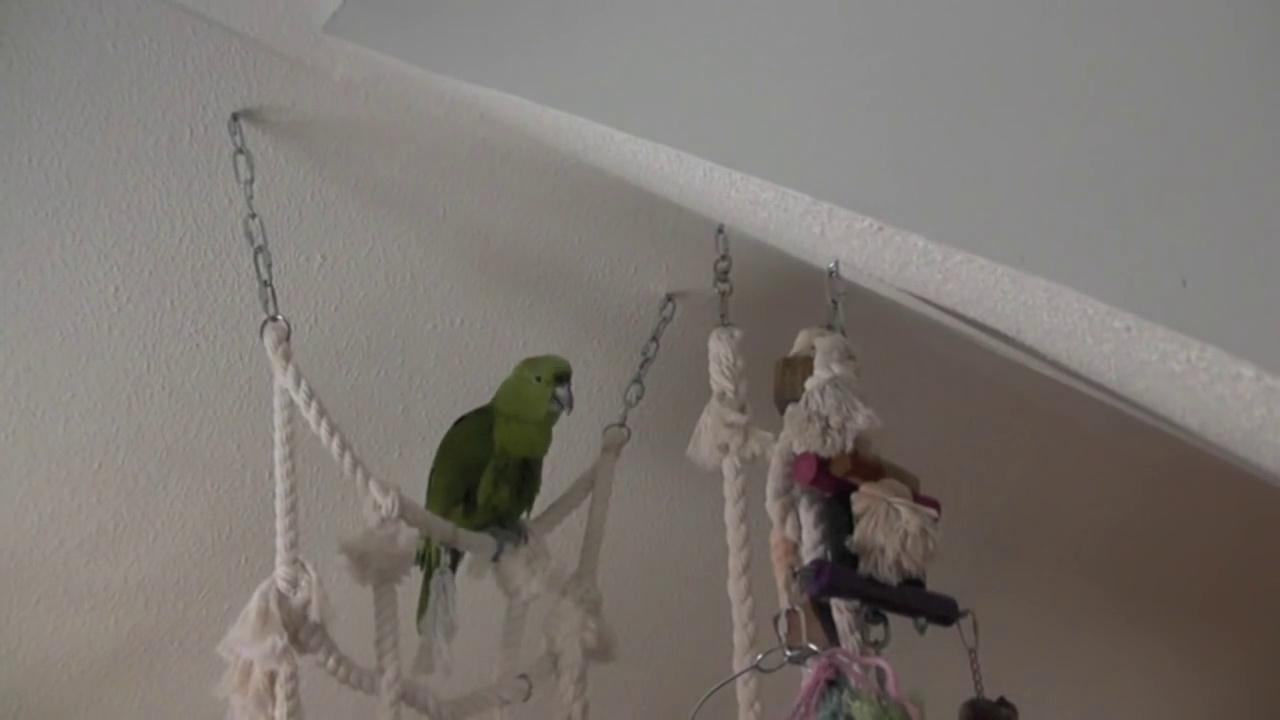}
		\label{fig:y equals x}
	\end{subfigure}
	\begin{subfigure}{0.09\textwidth}
		\centering
		\includegraphics[width=\textwidth]{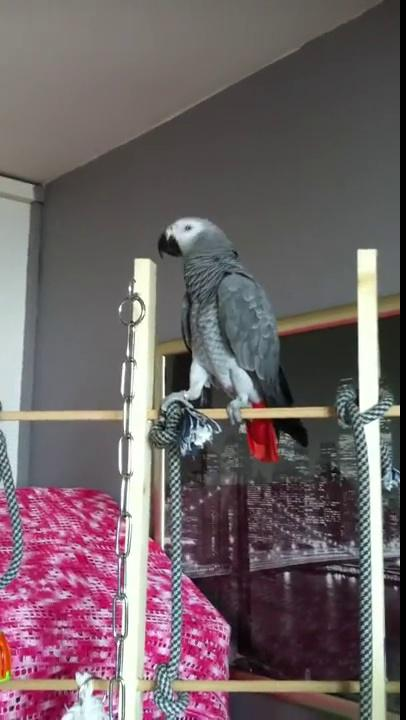}
		\label{fig:y equals x}
	\end{subfigure}
	\begin{subfigure}{0.09\textwidth}
		\centering
		\includegraphics[width=\textwidth]{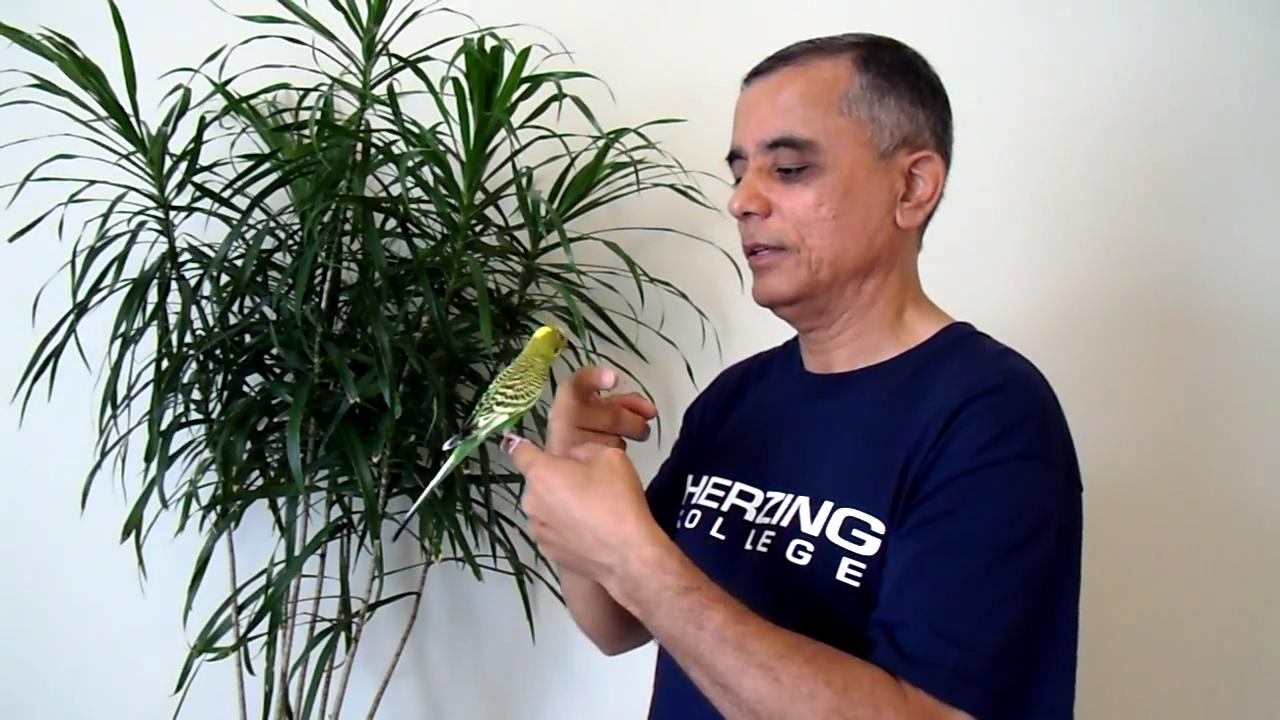}
		\label{fig:y equals x}
	\end{subfigure}
	\begin{subfigure}{0.09\textwidth}
		\centering
		\includegraphics[width=\textwidth]{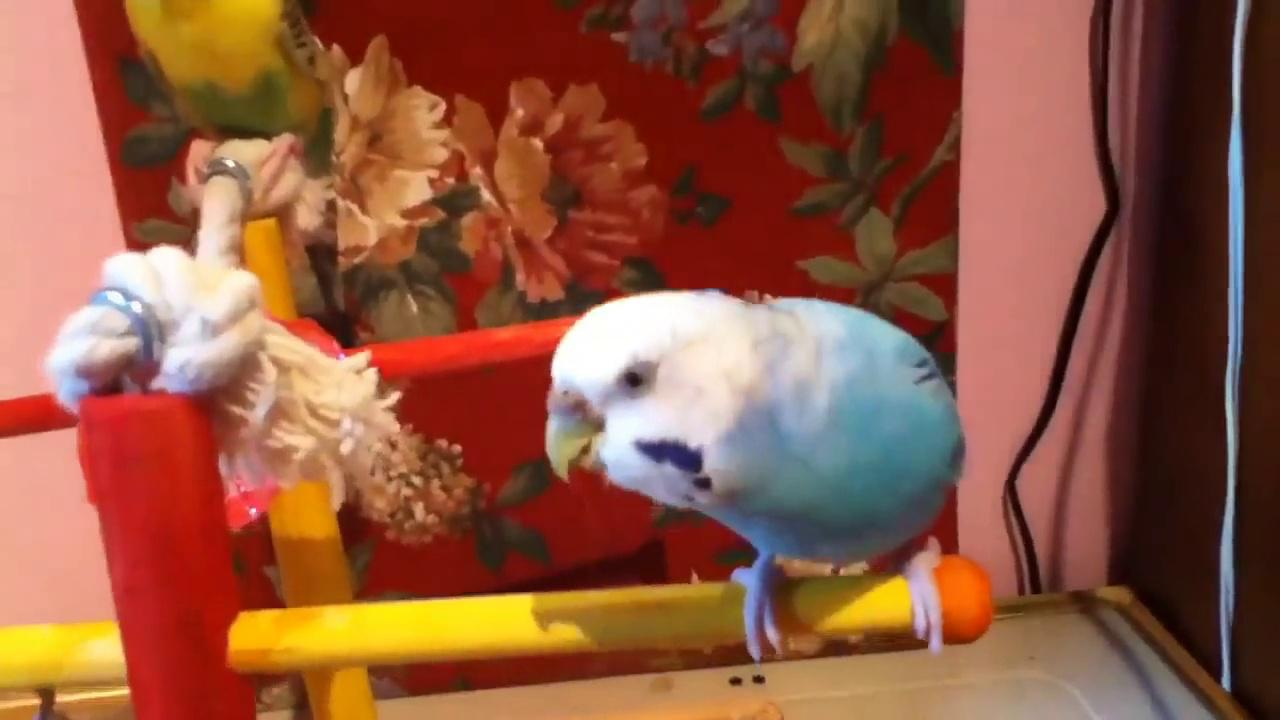}
		\label{fig:y equals x}
	\end{subfigure}
	\begin{subfigure}{0.09\textwidth}
		\centering
		\includegraphics[width=\textwidth]{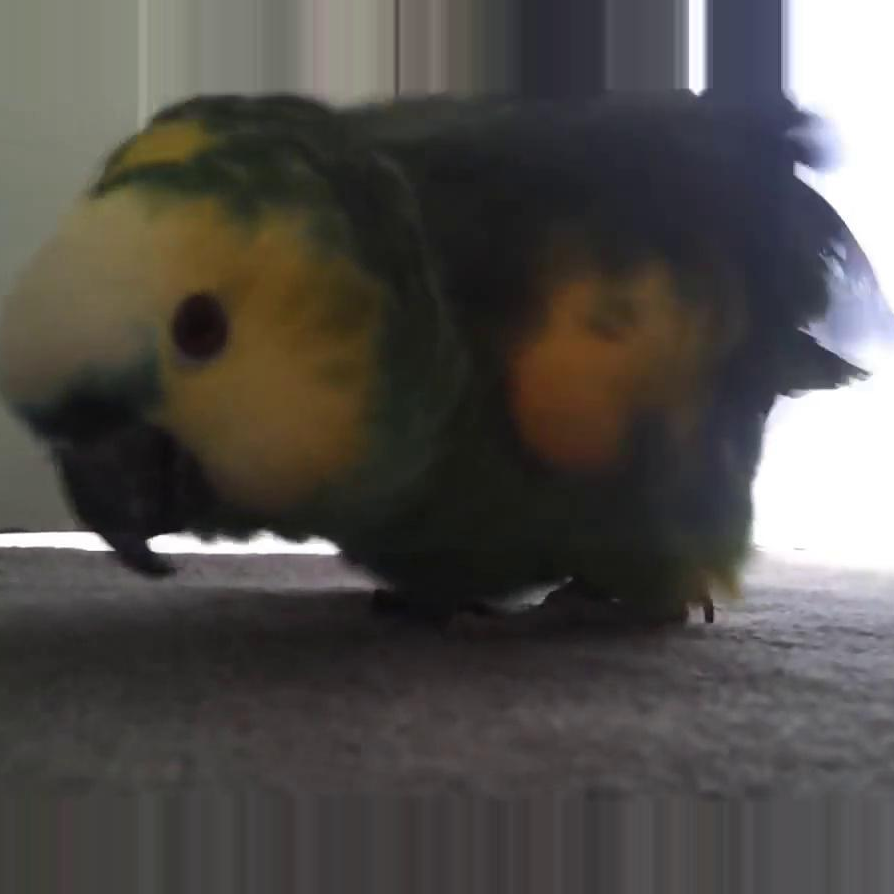}
		\label{fig:y equals x}
	\end{subfigure}
	\begin{subfigure}{0.09\textwidth}
		\centering
		\includegraphics[width=\textwidth]{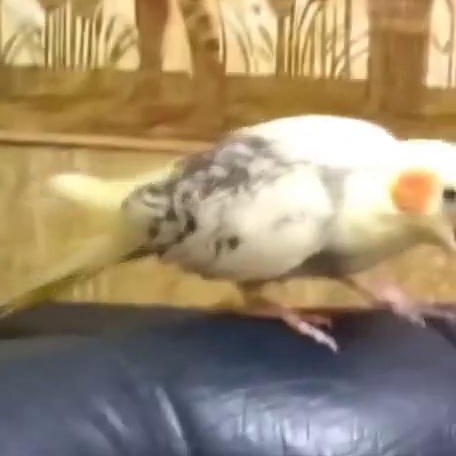}
		\label{fig:y equals x}
	\end{subfigure}
	\begin{subfigure}{0.09\textwidth}
		\centering
		\includegraphics[width=\textwidth]{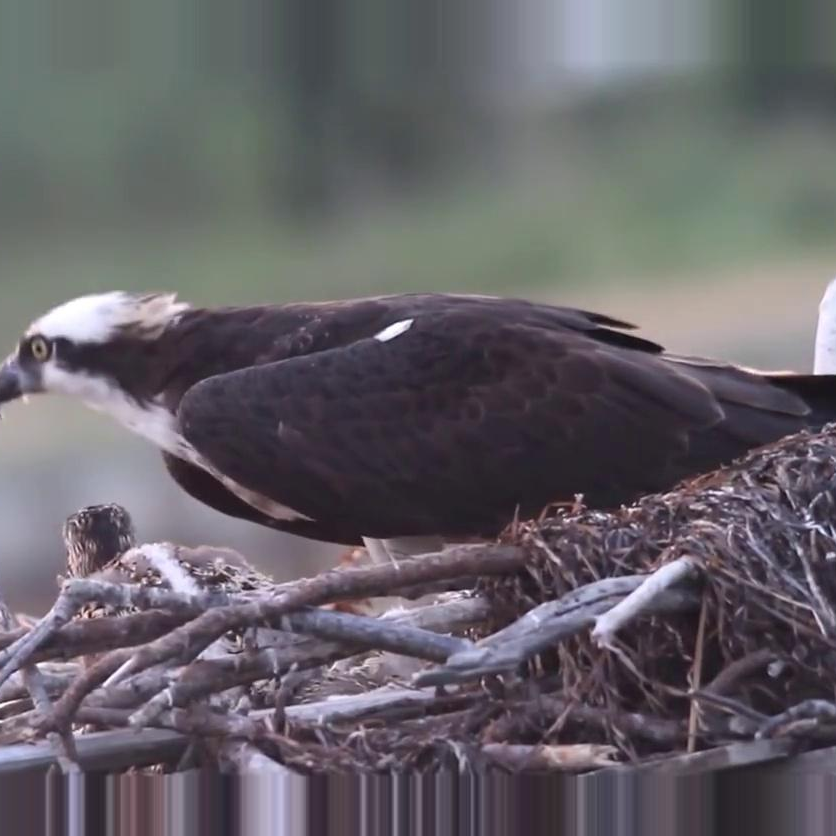}
		\label{fig:three sin x}
	\end{subfigure}
	\begin{subfigure}{0.09\textwidth}
		\centering
		\includegraphics[width=\textwidth]{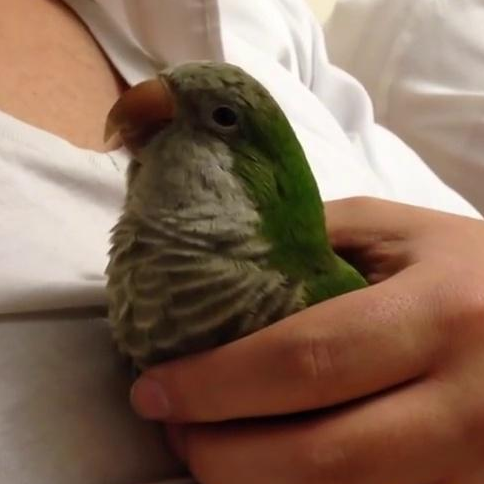}
		\label{fig:five over x}
	\end{subfigure}
	\vspace{-1em}
	\begin{subfigure}{0.09\textwidth}
		\centering
		\includegraphics[width=\textwidth]{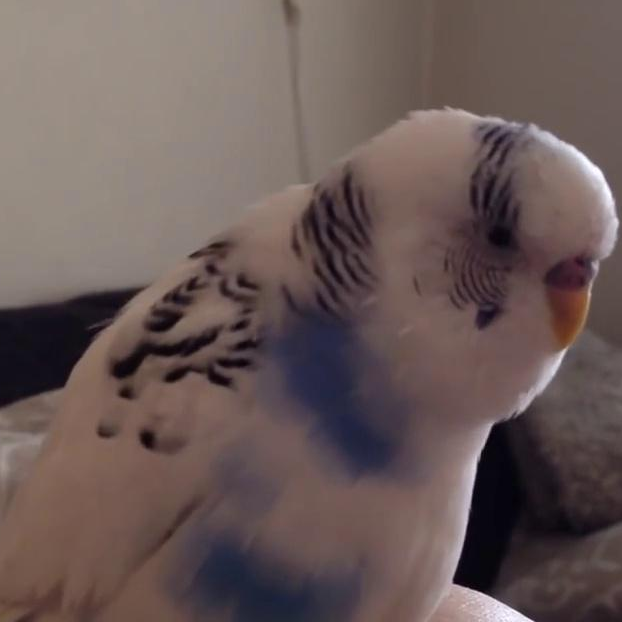}
		\label{fig:y equals x}
	\end{subfigure}
	\begin{subfigure}{0.09\textwidth}
		\centering
		\includegraphics[width=\textwidth]{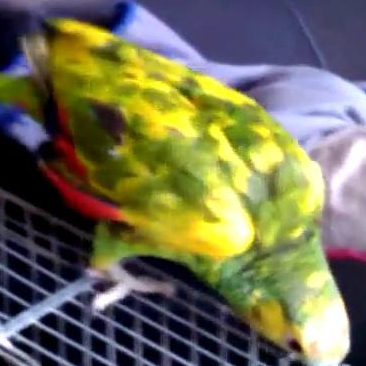}
		\label{fig:y equals x}
	\end{subfigure}
	\begin{subfigure}{0.09\textwidth}
		\centering
		\includegraphics[width=\textwidth]{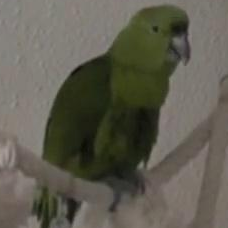}
		\label{fig:y equals x}
	\end{subfigure}
	\begin{subfigure}{0.09\textwidth}
		\centering
		\includegraphics[width=\textwidth]{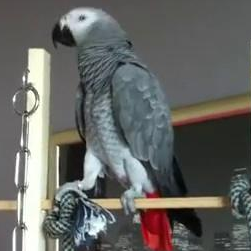}
		\label{fig:y equals x}
	\end{subfigure}
	\begin{subfigure}{0.09\textwidth}
		\centering
		\includegraphics[width=\textwidth]{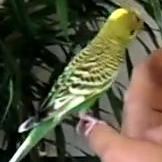}
		\label{fig:y equals x}
	\end{subfigure}
	\begin{subfigure}{0.09\textwidth}
		\centering
		\includegraphics[width=\textwidth]{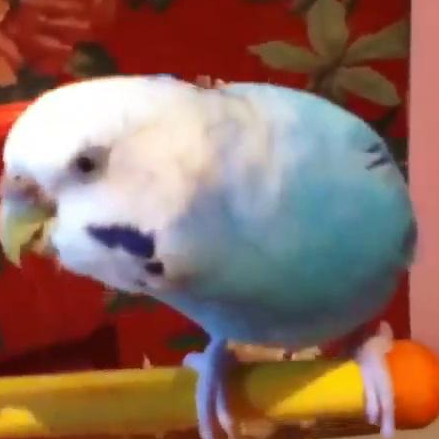}
		\label{fig:y equals x}
	\end{subfigure}
	\begin{subfigure}{0.09\textwidth}
		\centering
		\includegraphics[width=\textwidth]{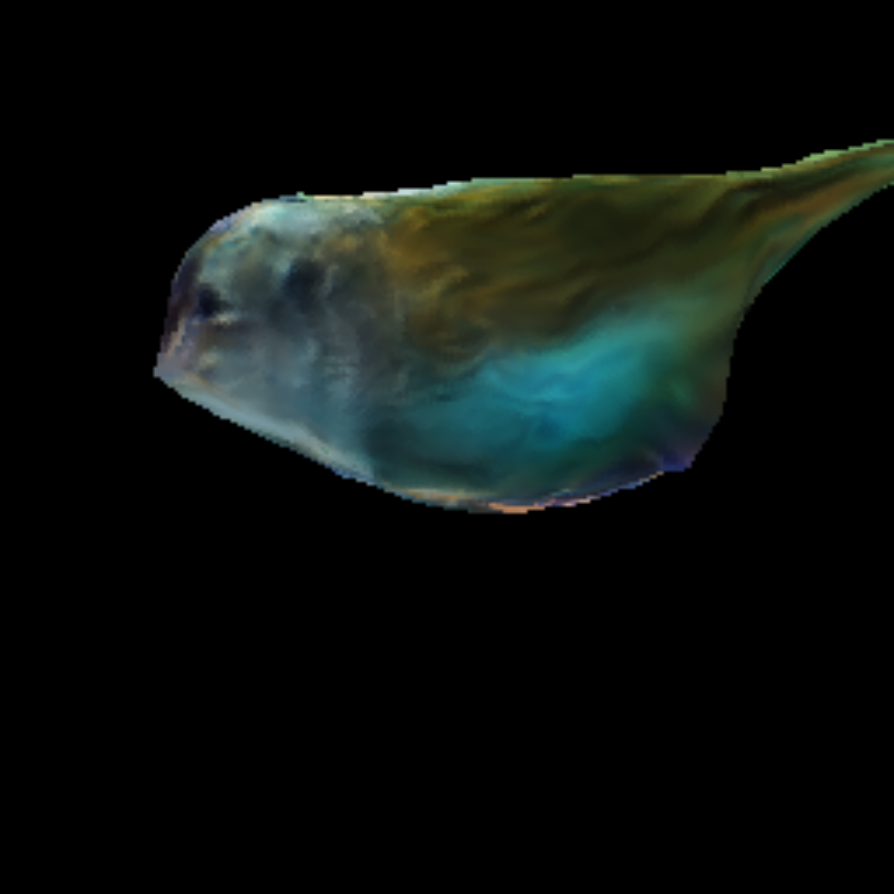}
		\label{fig:y equals x}
	\end{subfigure}
	\begin{subfigure}{0.09\textwidth}
		\centering
		\includegraphics[width=\textwidth]{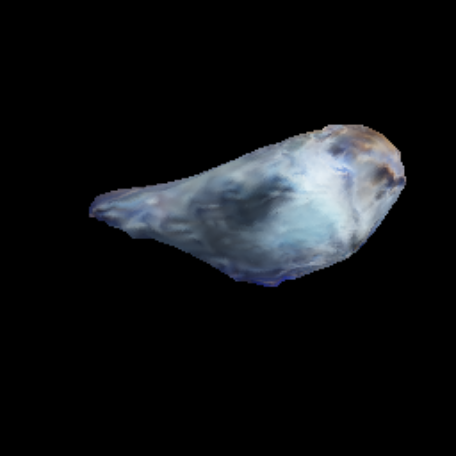}
		\label{fig:y equals x}
	\end{subfigure}
	\begin{subfigure}{0.09\textwidth}
		\centering
		\includegraphics[width=\textwidth]{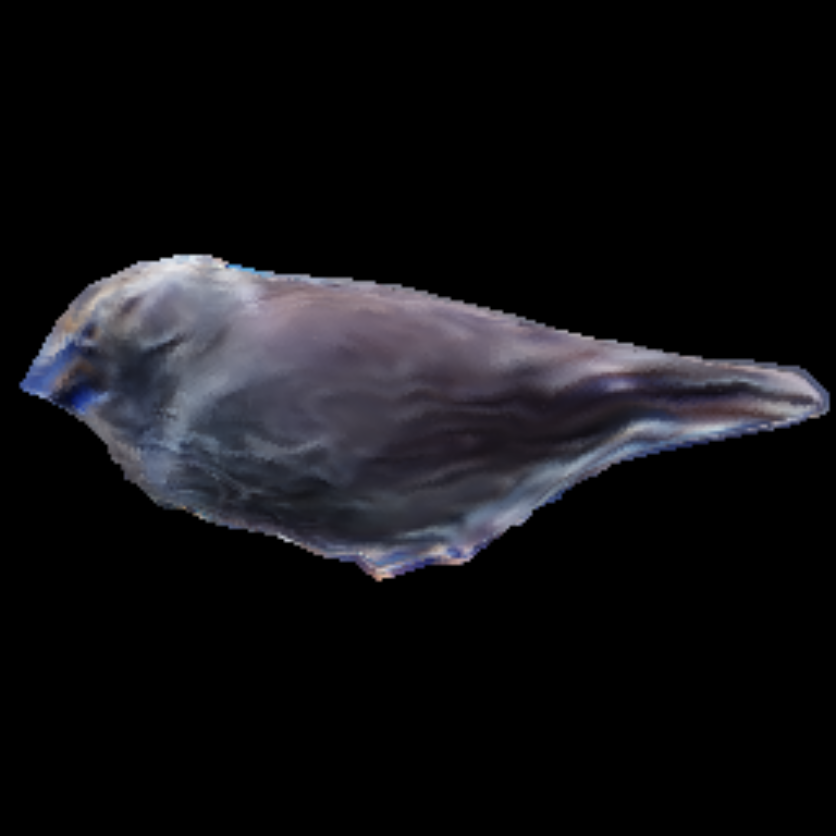}
		\label{fig:three sin x}
	\end{subfigure}
	\begin{subfigure}{0.09\textwidth}
		\centering
		\includegraphics[width=\textwidth]{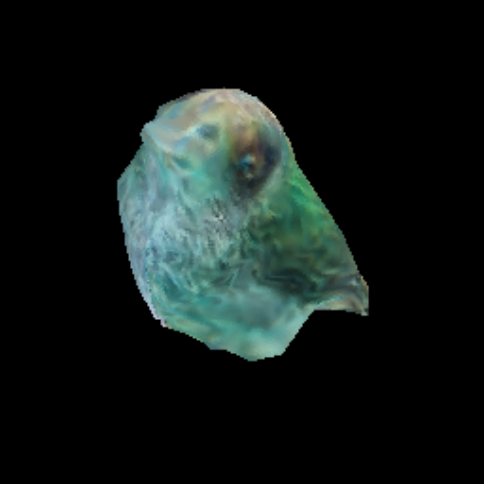}
		\label{fig:five over x}
	\end{subfigure}
	\vspace{-1em}
	\begin{subfigure}{0.09\textwidth}
		\centering
		\includegraphics[width=\textwidth]{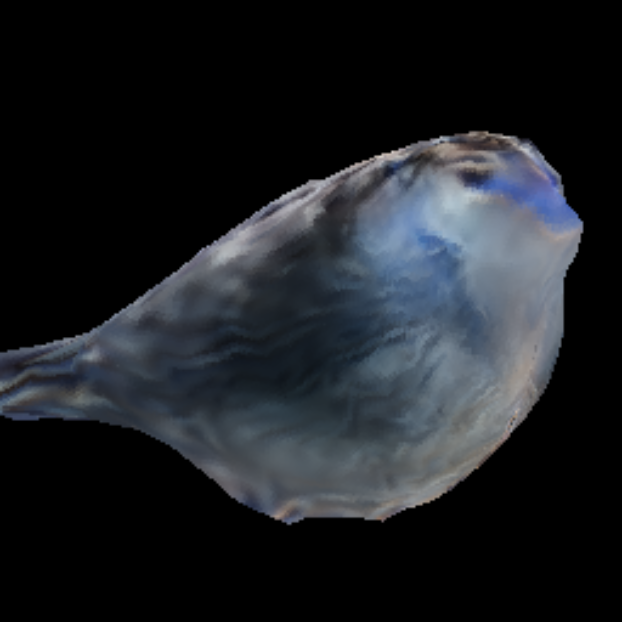}
		\label{fig:y equals x}
	\end{subfigure}
	\begin{subfigure}{0.09\textwidth}
		\centering
		\includegraphics[width=\textwidth]{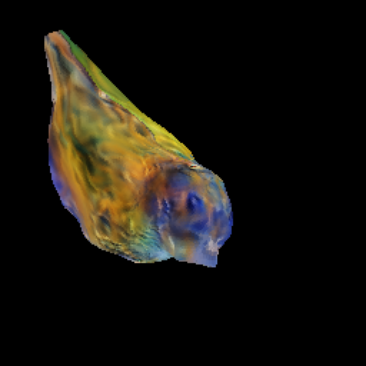}
		\label{fig:y equals x}
	\end{subfigure}
	\begin{subfigure}{0.09\textwidth}
		\centering
		\includegraphics[width=\textwidth]{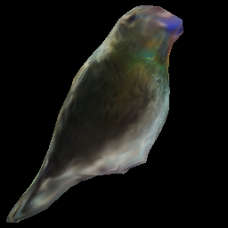}
		\label{fig:y equals x}
	\end{subfigure}
	\begin{subfigure}{0.09\textwidth}
		\centering
		\includegraphics[width=\textwidth]{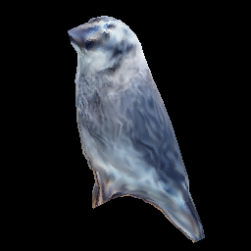}
		\label{fig:y equals x}
	\end{subfigure}
	\begin{subfigure}{0.09\textwidth}
		\centering
		\includegraphics[width=\textwidth]{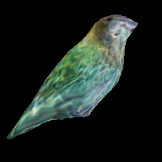}
		\label{fig:y equals x}
	\end{subfigure}
	\begin{subfigure}{0.09\textwidth}
		\centering
		\includegraphics[width=\textwidth]{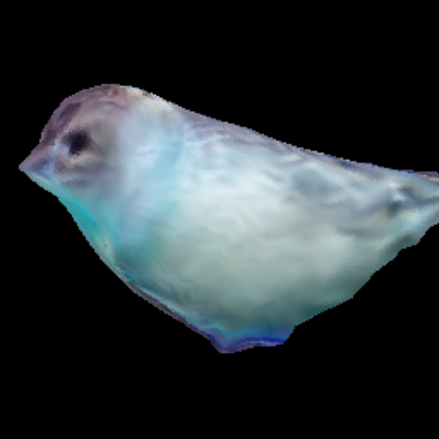}
		\label{fig:y equals x}
	\end{subfigure}
	\begin{subfigure}{0.09\textwidth}
		\centering
		\includegraphics[width=\textwidth]{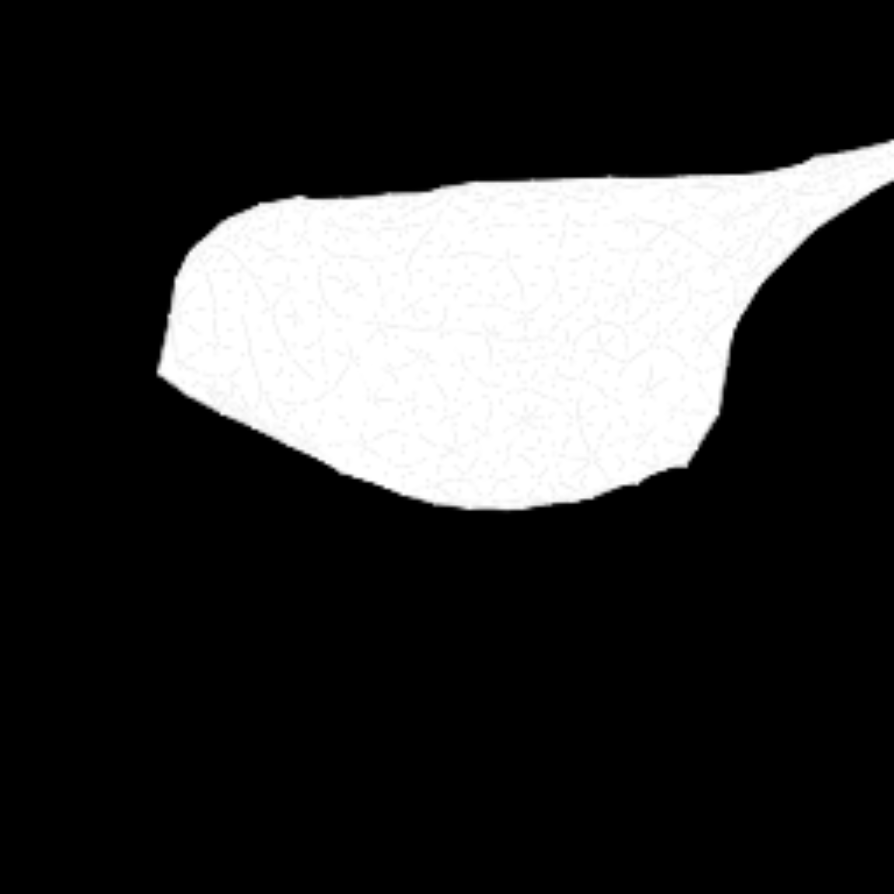}
		\label{fig:y equals x}
	\end{subfigure}
	\begin{subfigure}{0.09\textwidth}
		\centering
		\includegraphics[width=\textwidth]{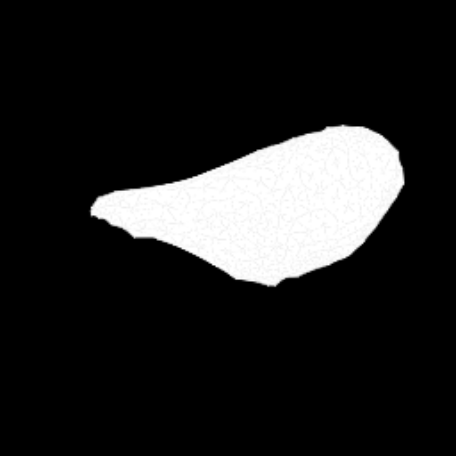}
		\label{fig:y equals x}
	\end{subfigure}
	\begin{subfigure}{0.09\textwidth}
		\centering
		\includegraphics[width=\textwidth]{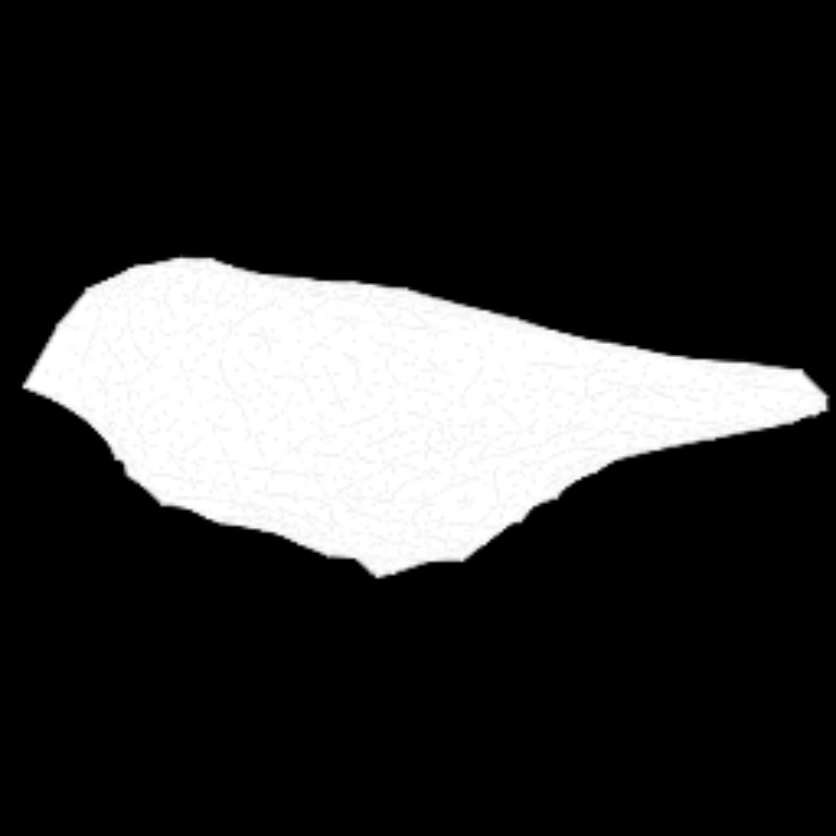}
		\label{fig:three sin x}
	\end{subfigure}
	\begin{subfigure}{0.09\textwidth}
		\centering
		\includegraphics[width=\textwidth]{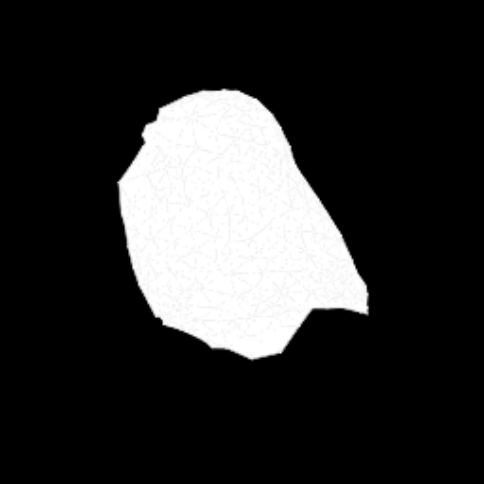}
		\label{fig:five over x}
	\end{subfigure}
	\vspace{-1em}
	\begin{subfigure}{0.09\textwidth}
		\centering
		\includegraphics[width=\textwidth]{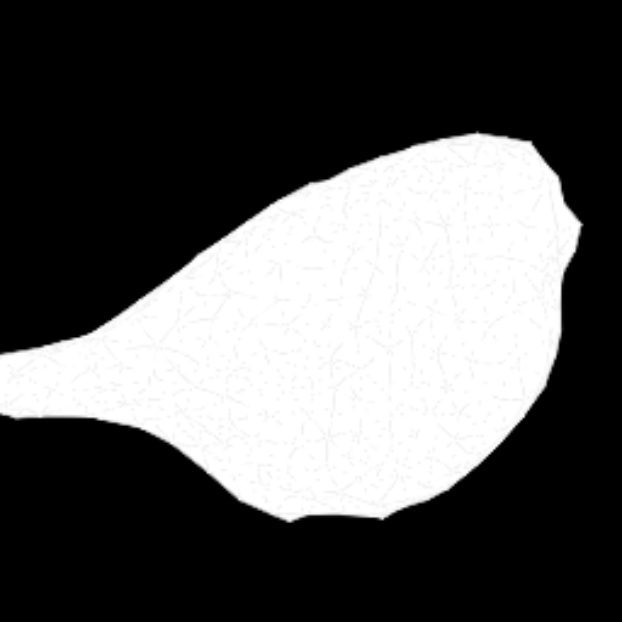}
		\label{fig:y equals x}
	\end{subfigure}
	\begin{subfigure}{0.09\textwidth}
		\centering
		\includegraphics[width=\textwidth]{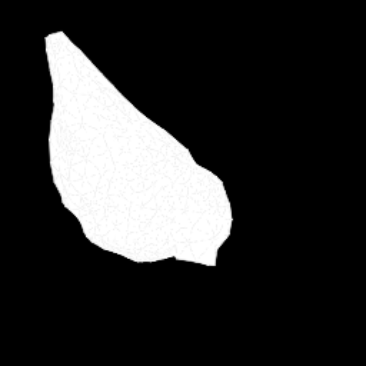}
		\label{fig:y equals x}
	\end{subfigure}
	\begin{subfigure}{0.09\textwidth}
		\centering
		\includegraphics[width=\textwidth]{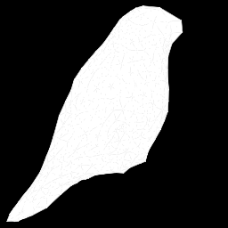}
		\label{fig:y equals x}
	\end{subfigure}
	\begin{subfigure}{0.09\textwidth}
		\centering
		\includegraphics[width=\textwidth]{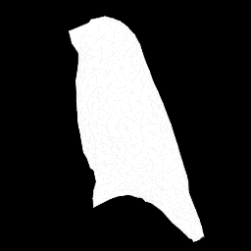}
		\label{fig:y equals x}
	\end{subfigure}
	\begin{subfigure}{0.09\textwidth}
		\centering
		\includegraphics[width=\textwidth]{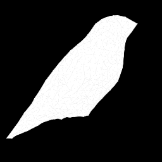}
		\label{fig:y equals x}
	\end{subfigure}
	\begin{subfigure}{0.09\textwidth}
		\centering
		\includegraphics[width=\textwidth]{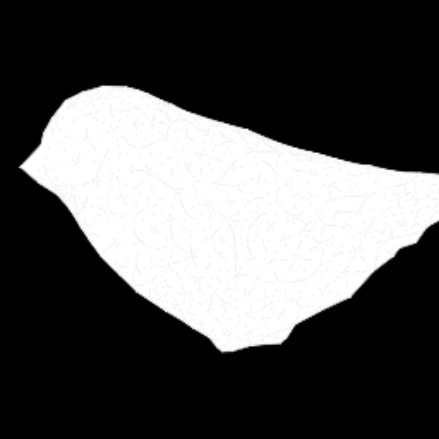}
		\label{fig:y equals x}
	\end{subfigure}
	\caption{On-line inference of 3D objects. The first row shows original images of 10 different birds of YouTubeVos and Davis \cite{xu2018youtube} video sequences. The second row presents the image patches extracted using width and height of the bounding boxes predicted by the LWL \cite{bhat2020learning} tracker. Third and forth rows show the reconstructed textures and masks using 3D object reconstruction model when keypoints correspondences is applied to predict camera poses capturing the images.}
	\label{fig:online}
\end{figure*}

\end{document}